\documentclass[a4paper,fleqn]{cas-dc}
\usepackage[authoryear]{natbib}

\usepackage{color}
\usepackage{float}
\usepackage{csquotes}
\usepackage{tabularx}

\usepackage[table]{xcolor}
\usepackage{colortbl}
\newcommand{\best}[1]{\cellcolor[gray]{0.85}\textbf{#1}}
\newcommand{\second}[1]{\cellcolor[gray]{0.93}#1}

\begin{document}


\let\WriteBookmarks\relax
\def\floatpagepagefraction{1}
\def\textpagefraction{.001}


\title [mode = title]{Auditing Patient Privacy in Medical Generative Models: Scalable Memorization Detection with DeepSSIM++}  
\shorttitle{DeepSSIM++} 
\shortauthors{Scardace et al.} 

\author[1]{Antonio Scardace}[orcid=0009-0006-7101-2879]
\ead{antonio.scardace@phd.unict.it}
\cormark[1]
\credit{Conceptualization, Methodology, Software, Validation, Formal Analysis, Investigation, Data Curation, Writing - Original Draft, Visualization}

\author[1]{Francesco Guarnera}[orcid=0000-0002-7703-3367]
\credit{Writing - Review \& Editing, Supervision}

\author[1]{Sebastiano Battiato}[orcid=0000-0001-6127-2470]
\credit{Writing - Review \& Editing, Supervision}

\author[2]{Daniele Ravì}[orcid=0000-0003-0372-2677]
\credit{Conceptualization, Writing - Review \& Editing, Project Administration, Supervision}


\affiliation[1]{organization={University of Catania},
    addressline={Department of Mathematics and Computer Science}, 
    city={Catania},
    country={Italy}
}

\affiliation[2]{organization={University of Messina},
    addressline={MIFT Department}, 
    city={Messina},
    country={Italy}
}


\cortext[1]{Corresponding author}


\begin{abstract}
While deep generative models offer new opportunities for medical image synthesis and data sharing, their ability to memorize and reproduce training samples raises serious concerns about patient confidentiality. Detecting such memorization at scale remains challenging: traditional pixel-based metrics are sensitive to generation artifacts, whereas generic embedding-based metrics often lack the anatomical sensitivity required for medical data. To address this challenge, we introduce DeepSSIM++, a self-supervised similarity metric for scalable memorization auditing in medical generative models. By leveraging multi-scale feature aggregation and anatomy-preserving augmentations, DeepSSIM++ learns an embedding space where cosine similarity approximates the Structural Similarity Index (SSIM), eliminating the need for exact pixel-level registration. Compared with state-of-the-art baselines, DeepSSIM++ achieves an average Macro F1 improvement of 33 percentage points under ideal alignment and 46 percentage points under realistic spatial and intensity perturbations. Furthermore, it accelerates large-scale similarity computation by several orders of magnitude compared with analytical SSIM. By combining anatomical sensitivity and computational efficiency, DeepSSIM++ provides an open-source tool for scalable memorization auditing in medical generative AI. Code and data are publicly available at: \url{https://github.com/brAIn-science/DeepSSIM}.
\end{abstract}





\maketitle


\section{Introduction}

Generative models are increasingly adopted in medical imaging, enabling applications such as image reconstruction~\citep{ahishakiye2021}, modality translation~\citep{sargood2026}, synthetic data generation for downstream clinical tasks such as disease progression modeling~\citep{brlp2025}, and data augmentation~\citep{Seo2025}. Among the most studied families of image generative models are Variational Autoencoders~\citep{vae2014}, Generative Adversarial Networks (GANs)~\citep{gan2014}, Diffusion Models~\citep{ddpm2020,ldm2022}, and Flow Matching models~\citep{lipman2023}. However, despite their remarkable success, a fundamental question remains: \textbf{do these models truly learn the underlying data distribution, or do they simply memorize their training data?} Ideally, a generative model should synthesize novel images that preserve the statistical properties of the training distribution without reproducing specific examples. When this property is violated and generated samples closely replicate training instances, the model is considered to exhibit memorization rather than generalization. The boundary between these two behaviors is governed by multiple interacting factors, including model architecture, training data characteristics, and optimization dynamics~\citep{bonnaire2026}.

This distinction becomes particularly important in privacy-sensitive applications, where synthetic data is often proposed as a privacy-preserving alternative to sharing real-world data. This assumption relies on the expectation that generative models capture the statistical properties of the training distribution without retaining information specific to individual training instances. For example,~\citet{jahanian2022} have suggested that diffusion models may improve privacy by generating synthetic samples rather than directly exposing real data. However, if a model memorizes and reproduces training examples, synthetic generation no longer guarantees privacy and may instead introduce risks of information leakage. The consequences are particularly severe in healthcare applications, where clinical data often contain fine-grained information that may act as subject-specific identifiers. Structural brain MRI scans, for example, feature highly detailed morphological patterns, such as cortical gyri or lesion boundaries, that may uniquely identify individual patients~\citep{chauvin2020}. In this context, memorization compromises patient confidentiality, reduces trust in AI systems, and slows the clinical adoption of otherwise promising generative AI technologies.

Reliable memorization assessment is therefore essential for comparing models and providing measurable evidence of privacy and reliability. Defining memorization, however, remains challenging because similarity can be evaluated at different levels. While natural images may exhibit semantic similarity despite substantial pixel-level differences, medical images require stricter criteria based on identity-preserving anatomical structures. In our work, we consider and extend the memorization definition introduced by~\citet{dar2023}, which was specifically designed for the medical imaging domain. Here, a generated sample is considered memorized if it closely reproduces a training instance up to minor affine transformations. Since our objective is to assess the risk of patient re-identification, we generalize this notion beyond affine transformations to include realistic artifacts and appearance distortions, such as intensity variations and noise. These perturbations may alter the visual appearance of a sample while preserving the subject-specific anatomical information that enables re-identification. Current approaches face significant challenges in identifying this type of memorization: they either rely on generic visual embeddings that overlook subtle anatomical similarities or require computationally expensive image comparisons that limit their scalability.

To address these limitations, we propose DeepSSIM++, a self-supervised metric designed to detect and quantify memorization in medical generative models. Existing pixel-based metrics often fail under minor spatial misalignment, and generic visual embeddings overlook subtle anatomical correspondence. Our approach is developed to bridge these gaps by learning a transformation-invariant structural feature space while enabling scalable memorization detection with low computational cost. We focus our empirical evaluation on 2D structural brain MRI, a domain representing a particularly critical scenario for patient confidentiality, as prior work has shown that the high individual specificity of anatomical morphology enables patient re-identification with 100\% accuracy~\citep{chauvin2020}.

More specifically, we extend our previous study~\citep{scardace2026} along four key directions:

\begin{itemize}
    \item We introduce a refined annotation protocol for the memorization benchmark, including revised labels and rigorous inter-rater agreement analysis.
    
    \item We enhance DeepSSIM through several architectural and optimization improvements, including multi-scale feature aggregation and Layer-wise Learning Rate Decay (LLRD)~\citep{bu2024}, leading to improved similarity estimation accuracy.
    
    \item We provide a rigorous validation of DeepSSIM++ through extensive ablation studies, explainability analysis using LayerCAM~\citep{jiang2021}.
    
    \item We demonstrate that DeepSSIM++ consistently outperforms existing memorization metrics and achieves orders-of-magnitude faster similarity computation compared with analytical SSIM, enabling scalable memorization auditing on large synthetic datasets.
\end{itemize}

The remainder of this paper is organized as follows: Section~\ref{sec:related_work} reviews related work on memorization and similarity metrics in generative modeling. Section~\ref{sec:methods} describes the proposed framework, the memorization taxonomy, and the construction of our benchmark dataset. Section~\ref{sec:experiments} presents the experimental setup together with a comprehensive quantitative, qualitative, and computational evaluation of the proposed approach. Finally, Section~\ref{sec:discussion} discusses the practical implications and limitations of our findings.


\section{Related Work} \label{sec:related_work}

Recent studies suggest that memorization is not an inherent property of a specific generative architecture. Instead, it emerges from the interplay of multiple factors, including the composition of the training data, the model architecture, and the optimization process.


\subsection{Memorization in Generative Models}

Early GAN literature linked memorization to mode collapse, where generators produce a restricted subset of samples rather than learning the full target distribution~\citep{arora2018,feng2021}. This reduced diversity can give a misleading impression of generalization while masking training-set memorization and identity leakage. These concerns become even more significant for diffusion models~\citep{yoon2023,bonnaire2026}, whose training objective differs fundamentally from that of GANs. Unlike GANs, which learn a generator that maps noise to data through adversarial training, diffusion models are trained using denoising score matching (DSM)~\citep{vincent2011}, which explicitly learns to reconstruct corrupted training instances. Recent theoretical analyses~\citep{gu2025} have shown that, with high model capacity, the optimal DSM solution can converge toward the empirical data distribution rather than the underlying population distribution, leading to significantly higher rates of training-data memorization. Empirical studies further support this concern, demonstrating that diffusion models can expose more training data than GANs~\citep{somepalli2023,carlini2023}.


\subsection{Memorization in Medical Generative Models}

Within the medical domain, empirical investigations indicate that Latent Diffusion Models (LDMs)~\citep{ldm2022} are highly susceptible to training set reproduction, memorizing up to 59\% of their training instances under specific conditions~\citep{dar2023}. More recently, large-scale multi-modal evaluations across diverse 2D and 3D imaging cohorts revealed that over 37\% of patient records could be identified as memorized, with nearly 70\% of generated samples corresponding to direct patient copies. Crucially, these findings demonstrate that memorization is not exclusive to DSM-based approaches, as alternative generative models were also found to reproduce training instances. However, their lower synthesis quality reduced the perceptual recognizability of these copies, suggesting that high generative fidelity plays a critical role in whether memorized content can be perceptually and radiologically identified. This finding was supported not only by automatic similarity measures but also by expert radiological assessment~\citep{dar2026}.


\subsection{What are the Major Causes of Memorization?} \label{sec:memo_reasons}

\paragraph{Training Data Properties.}

The statistical properties of the training dataset are a primary determinant of memorization behavior. Empirical evidence~\citep{feng2021,bonnaire2026,gu2025,dar2026} indicates that dataset size and diversity strongly influence memorization, as models trained on smaller or less variable cohorts are more likely to replicate training instances. Similarly, the presence of duplicated samples has been identified as a major trigger for memorization, since repeated exposure increases the tendency of generative models to reproduce training examples rather than generalize~\citep{somepalli2023,carlini2023}. \citet{yoon2023} demonstrated that lower-resolution data and structurally simpler distributions are more susceptible to memorization. Additionally, they show that severe class imbalance further exacerbates this behavior, making under-represented minority classes more likely to be memorized than majority classes.

\paragraph{Model Design and Capacity.}

The architecture and capacity of a generative model also influence its tendency to memorize training samples. In general, high-capacity models with a large number of parameters exhibit greater memorization potential, whereas under-parameterized models are more constrained to capture broader patterns in the data distribution~\citep{yoon2023,dar2026}. More detailed analyses~\citep{bonnaire2026,gu2025} have shown that increasing layer width is associated with increased memorization capacity. In contrast, increasing network depth exhibits a non-monotonic relationship: while adding layers initially enhances a model's ability to memorize, excessive depth may hinder optimization and reduce memorization rates. Beyond overall scale and depth, specific architectural components can also influence memorization behavior. Recent studies~\citep{gu2025,dar2026} have identified skip connections and conditioning mechanisms to significantly influence memorization. In particular, conditional diffusion models exhibit higher memorization rates than unconditional models, and even random or uninformative conditioning signals can increase the tendency to replicate training instances.

\paragraph{Training and Optimization Strategies.}

The optimization process also strongly influences when and how memorization occurs. \citet{bonnaire2026,gu2025,dar2026} have shown that models often undergo distinct learning phases: an early training stage characterized by high-quality generation, followed by a later stage in which memorization gradually emerges. Consequently, early stopping can act as an implicit regularization mechanism, whereas longer training schedules generally tend to increase memorization rates. Training hyperparameters can further modulate memorization behavior. In~\citet{gu2025}, larger batch sizes have been associated with increased memorization, whereas stronger regularization through weight decay can reduce memorization behavior. In medical generative models, training duration has also been identified as a key factor influencing memorization~\citep{dar2026}.


\subsection{Privacy Risks and Mitigation Strategies}

Memorization remains challenging to detect because models can generate visually convincing samples while still leaking training information, either by reproducing training examples directly or by preserving sensitive patterns learned during training~\citep{carlini2023,yoon2023}.

Such leakage may remain hidden during standard inference and only emerge through dedicated attacks. These include membership inference attacks, which aim to determine whether a specific sample was part of the training set~\citep{shokri2017,carlini2022}; model inversion approaches, which attempt to reconstruct sensitive information from model outputs or internal representations~\citep{fredrikson2015,zhang2020}; and data extraction attacks, which have demonstrated that memorized training examples can be recovered directly from generative models at scale~\citep{carlini2019,carlini2021,balle2021,nasr2025,chen2026}.

These findings have motivated the development of privacy-preserving training strategies aimed at reducing the amount of sensitive information retained by generative models. In particular, Differential Privacy~\citep{abadi2016,ziller2021} approaches aim to limit the amount of retained information by introducing controlled perturbations during optimization, whereas Federated Learning~\citep{mcmahan2017,sheller2020} approaches reduce the need for centralized data sharing by keeping patient data distributed across multiple institutions. However, evaluating whether such approaches effectively reduce memorization without destroying clinical utility remains challenging, making reliable, transformation-invariant quantification metrics essential for auditing privacy–utility trade-offs.


\subsection{How Can Memorization Be Quantified?}

Recent literature has proposed different memorization metrics that typically follow a two-step strategy: first, a similarity function is used to measure correspondence between generated and training images; second, similarity scores are converted into memorization decisions through predefined thresholds or ranking criteria.

\subsubsection{Embedding-Based Metrics}

Several approaches have been proposed to quantify memorization by combining image similarity measures with specific classification strategies. In~\citet{somepalli2023}, memorization is assessed by embedding both real and synthetic images using the pre-trained Self-Supervised for image Copy Detection (SSCD) model~\citep{pizzi2022}, and computing cosine similarities between the resulting embeddings. The resulting similarity scores are then converted into a memorization estimate by considering the 95th percentile of the similarity distribution as an indicator of the most similar generated-training pairs. In~\citet{chen2024}, the authors note that this strategy may underestimate memorization in heavy-tailed similarity distributions. More recently, \citet{chen2026} introduced two complementary metrics: the Average Memorization Score (AMS), which estimates the overall prevalence of memorization by measuring the average proportion of generated images that match training instances across predefined similarity thresholds, and the Unique Memorization Score (UMS), which quantifies the diversity of memorized samples by measuring how many distinct training examples are recovered from the generated set. Their framework computes similarity using normalized $\ell_2$ distance for low-resolution datasets, following~\citet{carlini2023}, and SSCD embeddings for high-resolution datasets, using two fixed thresholds to assign image pairs to predefined similarity categories. However, these approaches were primarily developed for natural images and rely on generic similarity measures, which may not capture the fine-grained anatomical structures required to assess memorization in medical imaging. In particular, SSCD is pre-trained on large-scale natural image datasets and is not explicitly designed to identify subtle domain-specific structures.

A related line of research approaches memorization as a duplication detection problem. For instance, SemDeDup~\citep{abbas2023} leverages embeddings from pre-trained foundation models, such as BioMedCLIP~\citep{zhang2024}, by first clustering samples in the embedding space and then identifying semantically duplicate pairs within each cluster based on a similarity threshold. Although originally proposed to improve data efficiency by removing redundant training samples, semantic deduplication frameworks can also be adapted to estimate memorization by identifying generated images that closely match training examples. However, general biomedical foundation models are typically optimized for high-level semantic tasks rather than fine-grained structural correspondence, making them less suitable for detecting strict anatomical duplication where preservation of individual structures is required.

\subsubsection{Domain-Specific Similarity Learning}

As an alternative,~\citet{dar2026} proposes training a feature extractor using contrastive learning directly on the same training set as the generative model under study. This approach avoids relying on external pre-trained representations, which may limit the detection of medically relevant memorization patterns. However, even domain-specific approaches based on learned representations may face challenges when detecting strict anatomical duplication.

\subsubsection{Pixel-Based Metrics}

Pixel-based metrics provide stronger structural correspondence but remain sensitive to spatial misalignment, causing direct pixel-space comparisons to fail under affine transformations~\citep{zhang2018}. This motivates the need for robust metrics capable of combining structural sensitivity with transformation invariance.


\subsection{Similarity Metrics in Medical Imaging}

Unlike natural image analysis, duplicate detection in medical images requires the precise preservation of anatomical structures, with near pixel-wise correspondence between images. To capture these clinically relevant details, the Structural Similarity Index (SSIM)~\citep{wang2004} is one of the standard similarity measures.

\citet{breger2025} examined SSIM's behavior by comparing original brain MRI scans with lower-quality reconstructions. Their results revealed that SSIM scores remained relatively stable even as image quality visibly deteriorated. While this insensitivity to degradation is a limitation for Image Quality Assessment tasks, it is highly advantageous for duplicate detection, enabling the identification of memorized anatomical features despite the generation artifacts. However, SSIM suffers from two critical bottlenecks: it assumes strict spatial alignment, which is often violated by generation artifacts, and its exhaustive pairwise comparisons become computationally prohibitive at scale.

Recent evidence further highlights the importance of robustness to image transformations for accurate memorization detection in realistic generation settings. \citet{dar2026} showed that embedding-based pipelines can identify memorized samples even when generated images undergo transformations such as rotations, flips, and contrast variations. While embedding-based methods provide the scalability required for large-scale evaluation and robustness to such perturbations, they often lack the structural sensitivity needed to capture subtle anatomical correspondences, motivating the development of specialized hybrid approaches.

These observations motivate DeepSSIM++, which combines the structural sensitivity required for medical image memorization analysis with the efficiency, scalability, and transformation robustness of learned representations, enabling large-scale evaluation of generative models.


\subsection{Legal and Regulatory Implications}

Beyond technical challenges, memorization in generative models raises significant legal, regulatory, and ethical concerns. Recent debates on copyright infringement in AI-generated images have questioned whether generative models unlawfully reproduce protected material~\citep{gillotte2020,butterick2023}. Similarly, privacy concerns have emerged regarding the unauthorized sharing of clinical data with proprietary models~\citep{agarwal2024}. These cases show that memorization extends beyond an academic issue, raising fundamental questions of ownership, consent, and legal compliance. The key concern is not only whether models generate novel content, but whether their outputs or training procedures reproduce protected information in ways that create liability.

In Europe, these concerns are increasingly reflected in regulatory frameworks such as the General Data Protection Regulation (GDPR)~\citep{eu2016gdpr} and the EU AI Act~\citep{eu2024aiact}, which establish requirements for transparency, risk management, and the trustworthy deployment of AI systems. In contrast, the United States regulatory framework, including the Health Insurance Portability and Accountability Act (HIPAA)~\citep{hipaaprivacyrule2003}, primarily focuses on the protection of individually identifiable health information. While the EU AI Act explicitly addresses emerging challenges introduced by AI technologies, HIPAA provides limited guidance on the specific risks associated with memorization in generative AI systems.


\section{Methods} \label{sec:methods}

\begin{figure*}[t]
\centering
\includegraphics[width=0.85\textwidth]{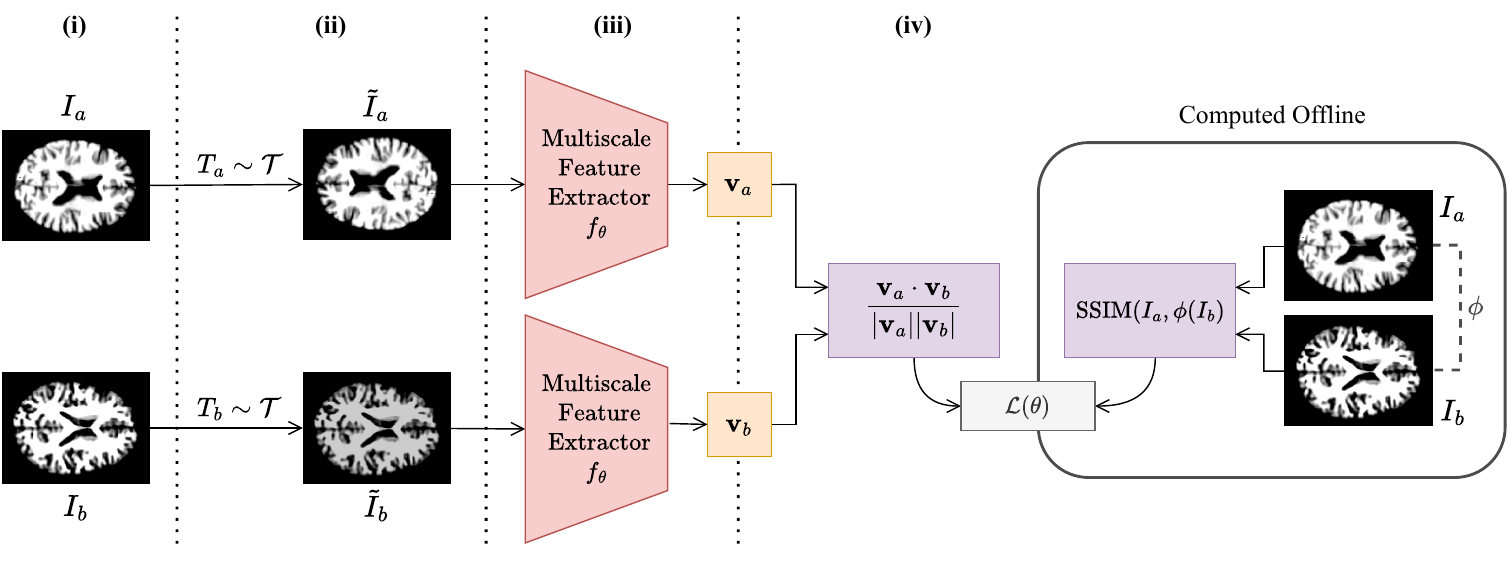}
\caption{ \label{fig:pipeline} \textbf{Training Pipeline:} The figure illustrates the DeepSSIM++ self-supervised training process, which is divided as follows: \textbf{(i)} Inputs: real and synthetic MRI scans are first preprocessed. \textbf{(ii)} Augmentation: random anatomical-preserving augmentations are applied before feature extraction to improve robustness to image variability. \textbf{(iii)} Embeddings: images are mapped into a lower-dimensional embedding space using a feature extractor $ f_\theta $. \textbf{(iv)} Model Optimization: the training loss minimizes the discrepancy between the embedding cosine similarity and the corresponding SSIM target, optimizing the feature extractor parameters $\theta$. }
\end{figure*}

In this section, we first introduce the datasets used to train and evaluate our framework. We then present the two main components of the proposed DeepSSIM++ pipeline for memorization detection. The first component, illustrated in Figure~\ref{fig:pipeline}, is a self-supervised model that learns a representation space where the cosine similarity between two image embeddings approximates the SSIM of the corresponding image pairs in the pixel domain. During training, SSIM targets are computed from spatially aligned image pairs to guide the learning of their structural similarity. The second component is a thresholding strategy to classify image pairs in this embedding space as \textit{Unrelated}, \textit{Near-Duplicates}, or \textit{Exact Duplicates} based on their similarity scores.


\subsection{Datasets}

We construct three datasets, each serving a specific role in our framework. The first dataset, denoted by $\mathcal{D}$, consists of real brain MRI scans and is used to train a standard LDM used to generate synthetic images under memorization-promoting conditions. The second dataset, denoted by $\mathcal{S}$, comprises the synthetic MRI slices generated by this LDM, which potentially contain memorized instances. The third dataset, denoted by $\mathcal{P}$, consists of real--synthetic image pairs with varying levels of similarity drawn from $\mathcal{D}$ and $\mathcal{S}$. This dataset forms the core focus of our analysis and is used to train and evaluate DeepSSIM++. 

We note that the specific generative architecture is not the subject of our investigation. In fact, the LDM considered here is utilized exclusively as a practical case study to demonstrate our primary contribution: a robust and scalable methodology for detecting memorization. Because our detection pipeline operates entirely on the generated images rather than internal model weights, our work provides a general framework suitable for evaluating any other medical generative architecture.

\subsubsection{Real MRI Dataset ($\mathcal{D}$)}

This dataset comprises 2,583 cross-sectional brain MRI scans from two public datasets: IXI (1,122 scans) and CoRR (1,461 scans). It includes 2,024 T1-weighted (T1w) and 559 T2-weighted (T2w) acquisitions. The data are randomly split into a training set $\mathcal{D}_{\mathrm{train}}$ (85\%) to train the LDM, and a validation set (15\%) used for early stopping.

\paragraph{MRI Preprocessing.} \label{sec:preproc}

All MRI scans in this dataset are processed using a standardized pipeline consisting of bias-field correction~\citep{Ants2021}, skull stripping~\citep{HdBet2019}, affine registration to MNI space~\citep{Ants2021}, and intensity normalization~\citep{WhiteStrip2014}. From each preprocessed 3D volume, we extract the central axial slice, resized to $112 \times 136$ pixels.

\subsubsection{Synthetic Dataset ($\mathcal{S}$)} \label{sec:memo_ldm}

The second dataset consists of synthetic MRI slices generated by the LDM. To increase the likelihood of generating memorized instances, we follow~\citet{dar2026}, who showed that using a synthetic set substantially larger than the real training set facilitates the creation of memorized samples. Specifically, for each real image present in $\mathcal{D}_{\mathrm{train}}$, we generate 30 synthetic samples, resulting in a synthetic dataset $\mathcal{S}$ containing 65,850 images. Additionally, to further increase training-data memorization, we incorporate three primary risk factors identified in~\citet{gu2025}: (i) controlled data duplication by randomly injecting 1--15 copies of each training image, (ii) an extended training schedule, and (iii) image-specific text conditioning, where each sample is associated with a unique textual description. Each prompt follows the template below:

\begin{displayquote}
    ``\textit{middle axial slice from a \texttt{<MRI SEQUENCE>} brain MRI (image ID \texttt{<MRI\_UID>}) acquired from a \texttt{<AGE>}-year-old \texttt{<BIOLOGICAL SEX>}, with subject ID \texttt{<SID>}, from the \texttt{<DATASET NAME>} dataset.}''
\end{displayquote}

The resulting text embeddings are extracted using PubMedBERT~\citep{gu2021} and incorporated into the LDM through a cross-attention mechanism. 

\begin{figure*}[t]
\centering
\includegraphics[width=0.8\textwidth]{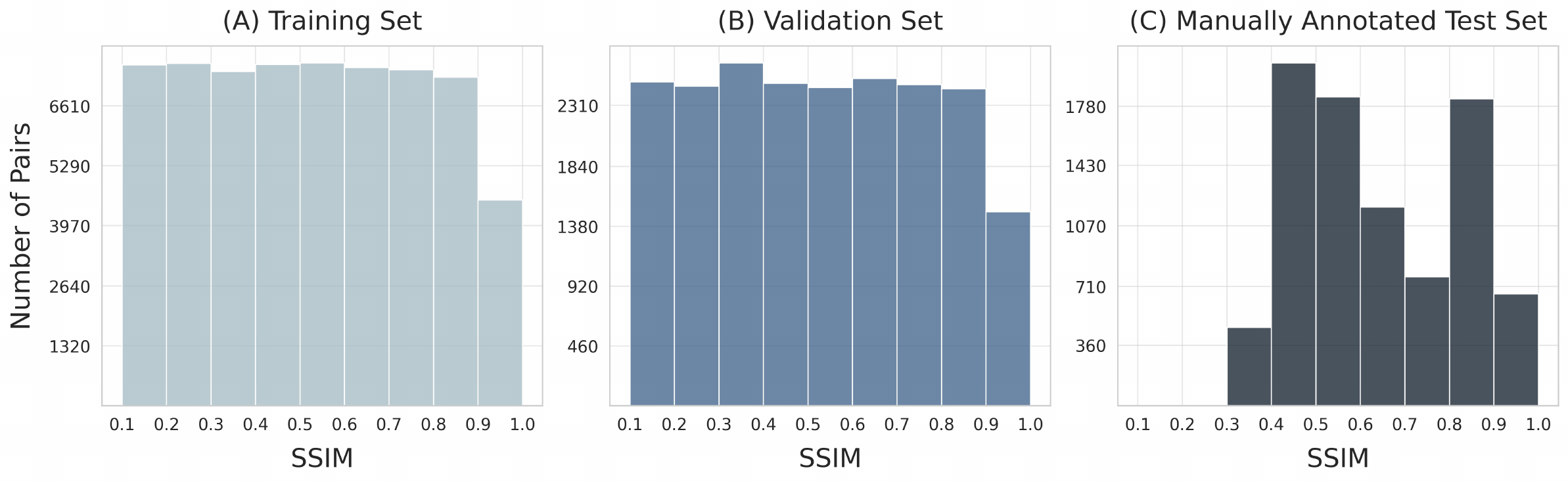}
\caption{ \label{fig:ssim_distr} \textbf{Distribution of Ground-Truth SSIM Scores.} The figure illustrates the distribution of image pairs across SSIM bins (width 0.1) for the three dataset splits. Panels \textbf{(A)} and \textbf{(B)} show the training and validation sets, $\mathcal{P}_{\mathrm{train}}$ and $\mathcal{P}_{\mathrm{valid}}$, respectively, both exhibiting approximately balanced distributions obtained by stratified random sampling. Panel \textbf{(C)} presents the manually annotated test set $\mathcal{P}_{\mathrm{test}}$, which is unbalanced due to the hybrid sampling strategy. }
\end{figure*}

\subsubsection{Real--Synthetic Pair Dataset ($\mathcal{P}$)} \label{sec:pairs_dataset}

The third dataset, denoted by $\mathcal{P}$, consists of real--synthetic image pairs. Specifically, each real image $I_a \in \mathcal{D}_{\mathrm{train}}$ is paired with every synthetic image $I_b \in \mathcal{S}$, resulting in $|\mathcal{D}_{\mathrm{train}}| \times |\mathcal{S}| = 2,195 \times 65,850 = 144,540,750$ pairs. Formally, this dataset is then defined as:

\begin{equation}
    \mathcal{P}=\{(I_a,I_b) \mid I_a\in\mathcal{D}_{\mathrm{train}},\, I_b\in\mathcal{S}\}.
\end{equation}

The dataset $\mathcal{P}$ is used to obtain three disjoint subsets: $\mathcal{P}_{\mathrm{train}}$, used to train our feature extractor $f_\theta$; $\mathcal{P}_{\mathrm{valid}}$, used for model selection of $f_\theta$ during the ablation study; and $\mathcal{P}_{\mathrm{test}}$, used for the final evaluation of the proposed model.

To prevent distributional biases during model training, which could skew DeepSSIM++ toward over-represented regions of the similarity spectrum, $\mathcal{P}_{\mathrm{train}}$ and $\mathcal{P}_{\mathrm{valid}}$ are obtained through stratified random sampling based on the reference SSIM scores, ensuring a balanced distribution of image pairs across the entire similarity range, as illustrated in Figure~\ref{fig:ssim_distr} (A) and (B). Additionally, we ensure that no image pair is assigned to more than one subset. 

For the test set $\mathcal{P}_{\mathrm{test}}$, applying the same SSIM-based sampling strategy would introduce an evaluation bias. Because the network is explicitly trained to predict SSIM values, using SSIM to curate the test set would evaluate the model on the same metric it was optimized to learn, potentially leading to overly optimistic performance estimates. To avoid this, we adopt a hybrid candidate selection strategy. For each real image, four image pairs are constructed: two using the synthetic samples with the highest FSIM scores~\citep{zhang2011fsim} and two using randomly selected synthetic samples. The FSIM-based selection, which relies on principles fundamentally different from SSIM, enriches the test set with highly similar cases, while random sampling preserves coverage of the natural similarity distribution. The resulting SSIM distribution of the test set obtained by this strategy is illustrated in Figure~\ref{fig:ssim_distr} (C).

The resulting test set contains 8,780 image pairs, independently annotated by three expert observers according to the similarity definitions proposed in Section~\ref{sec:labels}. Final labels are determined via majority voting, with no unresolved ambiguities across all annotated pairs. To assess annotation consistency, we measure the agreement among observers using Fleiss' $\kappa$, a statistical measure that evaluates how consistently observers assign the same labels while accounting for agreement that may occur by chance.



\subsection{Proposed Self-Supervised Similarity Metric}

We begin by computing SSIM-based supervision targets, which are used to train our model. For each pair $(I_a,I_b)$, where $I_a$ is a real image and $I_b$ is a synthetic image, we co-register $I_b$ to $I_a$ using a rigid transformation $\phi$. The aligned images are then normalized using z-score brightness normalization before computing the reference SSIM score:

\begin{equation}
    s_{ab}=\text{SSIM}(I_a,\phi(I_b)).
\end{equation}

Brightness normalization reduces the influence of non-anatomical intensity variations, ensuring that the target similarity primarily reflects structural correspondence. The co-registration step is used only during training to compute reliable SSIM targets. Because SSIM is highly sensitive to small spatial misalignments, aligning each image pair ensures that the supervision signal reflects structural similarity rather than positional differences. At inference time, no explicit co-registration is required, as the learned representation is designed to be robust to these variations.

To do so, we optimize a feature extractor $f_\theta$ that maps images into a representation space where the learned similarity score $s_\theta(I_a,I_b)$ approximates the corresponding SSIM target $s_{ab}$. Each image in the pair is then independently transformed by a randomly sampled transformation before feature extraction:

\begin{equation}
    \tilde{I}_a = T_a(I_a), \quad \tilde{I}_b = T_b(I_b), \quad T_a, T_b \sim \mathcal{T}, 
\end{equation}

where $\mathcal{T}$ denotes the set of training transformations. All transformations are designed to preserve anatomical identity while exposing the model to realistic spatial and appearance variations that may be encountered during inference. Consequently, the learned representation assigns high similarity scores to image pairs depicting the same anatomical content despite spatial misalignment or minor appearance distortions. The complete set of transformations used in our approach is described in Section~\ref{sec:augmentation}. The transformed images are then mapped into the representation space:

\begin{equation}
    \mathbf{v}_a = f_\theta(\tilde{I}_a), \quad \mathbf{v}_b = f_\theta(\tilde{I}_b).
\end{equation}

The feature extractor $f_\theta$ is optimized by minimizing the Mean Squared Error (MSE) between the learned similarity score $s_\theta$ and the SSIM supervision target over a batch of $B$ image pairs:

\begin{equation}
    \mathcal{L}(\theta)=
    \frac{1}{B}
    \sum_{i=1}^{B}
    \left(
    s_\theta(I_a^{(i)},I_b^{(i)})-s_{a_i b_i}
    \right)^2 .
\end{equation}

\subsubsection{Training Augmentations} \label{sec:augmentation}

To improve the robustness of DeepSSIM++ for memorization detection, we apply a set of anatomy-preserving augmentations during training. These transformations generate perturbed views while maintaining the underlying anatomical identity of each image. Formally, let $\mathcal{T} = \{T^{(1)}, \ldots, T^{(n)}\}$ denote the set of augmentation operators, each applied independently with probability $p^{(i)}$.

The augmentation strategy comprises two categories of transformations: intensity-based transformations, including bias-field augmentation, Gaussian noise, and contrast adjustment, which simulate appearance variations without modifying anatomical structures; and geometry-based transformations, including rotations, translations, zooming, and horizontal and vertical flips, which improve robustness to spatial perturbations and imperfect alignment. All transformations are designed to preserve anatomical correspondence between image pairs. Compared with the original DeepSSIM~\citet{scardace2026}, the proposed augmentation pipeline extends the set of transformations by introducing additional perturbations, including zooming, translations, bias-field augmentation, and Gaussian noise.

\begin{figure}[pos=t]
\centering
\includegraphics[width=0.9\linewidth]{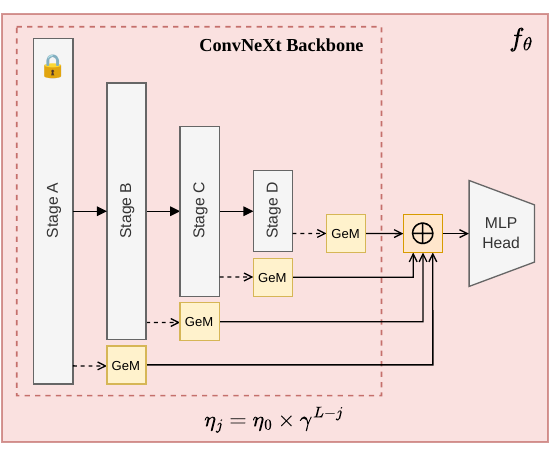}
\caption{ \label{fig:network} \textbf{Feature Extractor Architecture:} The figure illustrates the proposed feature extractor $f_\theta$, which combines a ConvNeXt backbone with multi-scale GeM pooling and an MLP-based fusion head. Feature maps are extracted from four progressively deeper stages (Stage A, B, C, and D), capturing increasingly abstract anatomical information. The stage outputs are independently pooled, concatenated, and projected into the final embedding space. }
\end{figure}

\subsubsection{Feature Extractor Architecture}

The network $f_\theta$, illustrated in Figure~\ref{fig:network}, combines a convolutional backbone with a multi-scale pooling and fusion head. We employ a pre-trained ConvNeXt-T backbone to extract feature maps from four progressively deeper stages of the network, denoted as Stages A, B, C, and D. As defined in the original architecture~\citep{liu2022}, these stages correspond to the main sequential blocks of the model, each operating at a progressively lower spatial resolution and higher feature dimensionality. Collectively, they capture hierarchical representations ranging from fine-grained anatomical details to high-level structural information~\citep{zeiler2014,wang2021}.

Unlike \citet{scardace2026}, which relies solely on the deepest representation, DeepSSIM++ aggregates feature maps from all four stages. Specifically, we propose that each stage output is independently pooled through a dedicated Generalized Mean (GeM) layer~\citep{radenovic2019} with a separately learned pooling exponent $p$. The resulting descriptors are concatenated and projected into the final embedding space through a two-layer MultiLayer Perceptron (MLP) comprising Layer Normalization, two linear layers separated by a GeLU activation, and dropout.

To account for the different adaptation requirements of the pre-trained backbone and the randomly initialized fusion head, we employ LLRD~\citep{bu2024}. This strategy assigns progressively smaller learning rates to earlier layers, preserving pre-trained representations while enabling the fusion head to adapt more rapidly. We treat the entire feature extractor as a sequence of $L$ trainable layers, and define the learning rate of layer $j$ as:

\begin{equation}
    \eta_j = \eta_0 \times \gamma^{L-j}.
\end{equation}

where $\eta_0$ is the base learning rate, $L$ is the total number of trainable layers, and $\gamma \in (0,1]$ is the decay factor.


\subsection{Proposed Memorization Metric}

For each pair $(I_a,I_b)$, where $I_a$ is a real image and $I_b$ is a synthetic image, we compute their structural similarity using the learned similarity function $s_\theta$. The resulting similarity score is converted into a semantic relationship between the two images using the thresholding function $\psi$:

\begin{equation}
    \psi(I_a, I_b) =
    \begin{cases}
        \text{Unrelated} & \text{if } s_\theta(I_a,I_b) \leq \alpha \\
        \text{Near-Duplicate} & \text{if } \alpha < s_\theta(I_a,I_b) < \beta \\
        \text{Exact Duplicate} & \text{if } s_\theta(I_a,I_b) \geq \beta
    \end{cases},
\end{equation}

where $\alpha$ and $\beta$ denote the decision thresholds and are hyperparameters of the proposed pipeline. 

\subsubsection{Definition of Memorization Categories} \label{sec:labels}

The three mutually exclusive categories defined by $\psi$ describe different levels of anatomical correspondence between a training image and a generated sample.

\paragraph{Exact Duplicate.} A pair $(I_a, I_b)$ is classified as an \textit{Exact Duplicate} if and only if every anatomical structure present in $I_a$ has a corresponding structure in $I_b$, and no structural deviation is observed between corresponding structures. Only anatomy-independent artifacts that do not alter the position, shape, size, or boundary of any individual anatomical structure are tolerated, such as uniform blur.

\paragraph{Near-Duplicate.} A pair $(I_a, I_b)$ is classified as a \textit{Near-Duplicate} when the overall anatomical configuration of $I_b$ is still recognizable as corresponding to $I_a$, but one or more localized deviations are present. These deviations are confined to specific anatomical regions and may involve boundary distortions, shape alterations, or tissue composition inconsistencies, without compromising the identifiability of the underlying anatomical correspondence.

\paragraph{Unrelated.} A pair $(I_a, I_b)$ is classified as \textit{Unrelated} when the anatomical structure of $I_b$ is not consistent with that of $I_a$, and no meaningful anatomical correspondence can be established between the two images.


\subsection{Implementation Details}

The framework is implemented in PyTorch~\citep{paszke2019}, leveraging MONAI~\citep{cardoso2022} for anatomy-preserving data augmentations. The feature extractor $f_\theta$ is based on a ConvNeXt-T architecture~\citep{liu2022} pre-trained on ImageNet. The original classification layer is replaced with an identity mapping, enabling direct extraction of image embeddings rather than image classification. To preserve low-level spatial features while adapting the model to the neuroimaging domain, all parameters in Stage A (including the stem and the initial convolutional layers) are frozen, whereas the remaining backbone stages and the projection head are jointly fine-tuned.

The network is trained for $65$ epochs, empirically determined via validation convergence, using the AdamW optimizer~\citep{loshchilov2019} with a weight decay of $1 \times 10^{-3}$ and a batch size of $64$. We employ a base learning rate $\eta_0 = 1 \times 10^{-3}$ for the projection head, which is then scaled across the trainable backbone layers using LLRD with a decay factor $\gamma=0.3$. The multi-scale GeM pooling layers are initialized with $p=3.0$ and $\epsilon=10^{-6}$, and the projection head uses a dropout probability of $0.1$ to produce final embeddings of dimension $d=256$.

The thresholds of the decision function $\psi$ are selected via grid search over $[0,1]$ with a step size of $0.01$, such that $\alpha < \beta$, resulting in $\alpha=0.76$ and $\beta=0.92$.

The LDM used as a case study is trained following the protocol proposed by~\citet{pinaya2022}, adapted to generate 2D axial brain MRI slices.

All model training procedures and experiments were performed on a workstation equipped with an Intel(R) Core(TM) i7-14700 CPU, $32$ GB of RAM, and a single NVIDIA RTX 4070 Ti SUPER GPU (16 GB of VRAM).


\section{Experiments and Results} \label{sec:experiments}

In this section, we conduct an extensive evaluation of DeepSSIM++ through five experiments: \textbf{(i)} an ablation study to identify the optimal configuration and quantify the impact of each design choice; \textbf{(ii)} a quantitative comparison of DeepSSIM++ for memorization detection against state-of-the-art baselines; \textbf{(iii)} a qualitative and explainability analysis to investigate the learned similarity representations; \textbf{(iv)} a sensitivity analysis of the proposed classification thresholds; and \textbf{(v)} a computational efficiency analysis comparing DeepSSIM++ with SSIM.

All experiments were performed using fixed random seeds to ensure reproducibility across data splitting, network initialization, and optimization.


\begin{table}[pos=t]
\centering
\small
\setlength{\tabcolsep}{3pt}
\def\arraystretch{1.15}

\begin{tabular}{lcc}
    \toprule
    \textbf{Experiment Setting} & \textbf{MAE} & \textbf{RMSE} \\

    \midrule
    \multicolumn{3}{c}{(A) Ablation on the Augmentations} \\
    \midrule
    Base Augmentations & $0.0474 \pm 0.0461$ & $0.0661 \pm 0.0129$ \\
    \best{Enriched Augmentations} & \best{$0.0452 \pm 0.0433$} & \best{$0.0626 \pm 0.0103$} \\
    
    \midrule
    \multicolumn{3}{c}{(B) Ablation on the Architecture} \\
    \midrule
    Single-Scale & $0.0452 \pm 0.0433$ & $0.0626 \pm 0.0103$ \\
    \best{Multi-Scale} & \best{$0.0361 \pm 0.0369$} & \best{$0.0517 \pm 0.0082$} \\
 
    \midrule
    \multicolumn{3}{c}{(C) Ablation on the Training Strategy} \\
    \midrule
    From Scratch & $0.0607 \pm 0.0497$ & $0.0784 \pm 0.0107$ \\
    Full Fine-Tuning & $0.0521 \pm 0.0481$ & $0.0709 \pm 0.0125$ \\
    \best{Partial Fine-Tuning} & \best{$0.0361 \pm 0.0369$} & \best{$0.0517 \pm 0.0082$} \\
    Head-Only Training & $0.0439 \pm 0.0446$ & $0.0626 \pm 0.0097$ \\

    \midrule
    \multicolumn{3}{c}{(D) Ablation on the Optimization Scheme} \\
    \midrule
    Uniform & $0.0361 \pm 0.0369$ & $0.0517 \pm 0.0082$ \\
    \best{LLRD} & \best{$0.0326 \pm 0.0383$} & \best{$0.0503 \pm 0.0090$} \\

    \midrule
    \multicolumn{3}{c}{(E) Ablation on the Loss Function} \\
    \midrule
    \best{MSE} & \best{$0.0326 \pm 0.0383$} & \best{$0.0503 \pm 0.0090$} \\
    MAE & $0.0335 \pm 0.0435$ & $0.0549 \pm 0.0120$ \\
    Soft Contrastive Loss & $0.0497 \pm 0.0395$ & $0.0635 \pm 0.0100$ \\

    \midrule
    \multicolumn{3}{c}{(F) Ablation on the Embedding Dimension} \\
    \midrule
    64 & $0.0326 \pm 0.0386$ & $0.0505 \pm 0.0093$ \\
    128 & $0.0326 \pm 0.0392$ & $0.0504 \pm 0.0092$ \\
    \best{256} & \best{$0.0325 \pm 0.0383$ }& \best{$0.0503 \pm 0.0090$} \\
    512 & $0.0338 \pm 0.0381$ & $0.0509 \pm 0.0089$ \\

    \midrule
    \multicolumn{3}{c}{(G) Ablation on the Backbone Architecture} \\
    \midrule
    ResNet50 & $0.0458 \pm 0.0440$ & $0.0635 \pm 0.0108$ \\
    DenseNet161 & $0.0603 \pm 0.0547$ & $0.0814 \pm 0.0144$ \\
    \best{ConvNeXt-T} & \best{$0.0325 \pm 0.0370$} & \best{$0.0492 \pm 0.0081$} \\

    \midrule
    \multicolumn{3}{c}{(H) Ablation on Hyperparameters} \\
    \midrule
    Default & $0.0325 \pm 0.0370$ & $0.0492 \pm 0.0081$ \\
    \best{Optimized} & \best{$0.0269 \pm 0.0310$} & \best{$0.0410 \pm 0.0056$} \\
    
    \bottomrule
\end{tabular}

\caption{ \label{tab:ablation} Results of the sequential ablation study on the validation set. Performance is reported in terms of MAE ($\pm$ SD) and RMSE ($\pm$ SD), evaluating the model’s ability to accurately predict SSIM values. At each step, the best-performing configuration identified in the previous evaluation is retained for subsequent experiments (highlighted in bold with a gray background). }
\end{table}


\subsection{Ablation Study}

To establish the optimal DeepSSIM++ configuration, we conducted an ablation study covering architectural choices, training strategies, optimization schemes, and hyperparameter tuning. All experiments were performed on the validation set $\mathcal{P}_{\mathrm{valid}}$, and performance was measured in terms of Mean Absolute Error (MAE) and Root Mean Squared Error (RMSE) between the predicted similarity scores from DeepSSIM++ and the reference SSIM values for each pair.

We adopt a sequential ablation strategy, evaluating one design choice at a time while retaining the best-performing configuration at each step. The base configuration follows the original DeepSSIM design described in~\citet{scardace2026}, with a ConvNeXt-B backbone, 256-dimensional output, partial fine-tuning of the pre-trained backbone, and MSE as the training objective.

\paragraph{Augmentations.} 

We first evaluate the impact of the enriched augmentation strategy defined in Section~\ref{sec:augmentation}. As shown in Table~\ref{tab:ablation}, expanding the set of transformations improves over the base configuration, reducing MAE by 4.64\% and RMSE by 5.30\%. This suggests that exposing the network to a wider range of realistic spatial and intensity perturbations improves robustness to such variations when learning the similarity metric.

\paragraph{Architecture.} 

Next, we compare the single-scale baseline---which extracts features exclusively from the deepest network stage---with the proposed multi-scale architecture. As shown in Table~\ref{tab:ablation}, the multi-scale variant achieves a reduction in MAE of 20.13\% and in RMSE of 17.41\%. These results suggest that aggregating hierarchical spatial representations at different granularities consistently improves the approximation of structural similarity.

\paragraph{Training Strategy.} 

We evaluate four training strategies to investigate how different fine-tuning schemes affect the adaptation of the backbone to the similarity prediction task: (i) training the network from scratch; (ii) head-only training, where the backbone is frozen and only the prediction head is trained; (iii) full fine-tuning, updating the entire backbone; and (iv) partial fine-tuning, selectively adapting a subset of pre-trained layers. The results in Table~\ref{tab:ablation} reveal a clear performance ordering that highlights the trade-off between preserving pre-trained visual knowledge and adapting to the target domain. Partial fine-tuning achieves the best performance, reducing MAE by 17.77\% and RMSE by 17.41\% compared with the second-best configuration (head-only training). Full fine-tuning produces a performance degradation, increasing MAE by 44.32\% relative to partial fine-tuning, suggesting that updating the entire backbone leads to excessive adaptation and disrupts the pre-trained representations. Finally, training the network from scratch yields the worst performance across all configurations, increasing MAE by 68.14\% relative to partial fine-tuning, further underscoring the critical importance of transfer learning from pre-training on natural images. Overall, these results indicate that freezing too much of the backbone limits adaptation, whereas updating too much of it disrupts transferable representations; consequently, selectively adapting a subset of pre-trained layers provides the optimal balance.

\paragraph{Optimization Scheme.} 

We evaluate the impact of the optimization strategy by comparing LLRD with a uniform learning rate strategy. In Table~\ref{tab:ablation}, we can see that LLRD consistently achieves better performance by assigning progressively smaller learning rates to earlier backbone layers and larger learning rates to the prediction head, enabling a more controlled adaptation of the pre-trained representations. Compared with the uniform optimization strategy, LLRD reduces MAE by 9.70\% and RMSE by 2.71\%, confirming the benefit of layer-wise adaptation for this task.

\paragraph{Loss Function.} 

We evaluate MSE, MAE, and Soft Contrastive Loss as supervision objectives for similarity learning. In Table~\ref{tab:ablation}, MSE achieves the best performance, reducing MAE by 2.69\% and RMSE by 8.38\% compared with the second-best loss function (MAE). This result suggests that the quadratic penalty provides a more effective optimization objective for this regression problem, as it places greater emphasis on larger prediction errors. Conversely, the contrastive loss underperforms in this direct regression setting, likely because its objective is not specifically designed to regress continuous similarity scores.

\paragraph{Embedding Dimension.} 

We evaluate four embedding dimensions corresponding to the backbone output dimensionality: 64, 128, 256, and 512. As shown in Table~\ref{tab:ablation}, an embedding dimension of 256 achieves the best overall performance. However, the average relative deviation across all dimensions is 1.54\% for MAE and 0.60\% for RMSE, suggesting that DeepSSIM++ is robust to this hyperparameter.

\paragraph{Backbone Architecture.}

We evaluate different convolutional backbones to assess their impact on similarity prediction, including ResNet~\citep{he2016}, DenseNet~\citep{huang2017}, and ConvNeXt~\citep{liu2022} variants with different capacities. As shown in Table~\ref{tab:ablation}, ConvNeXt-T achieves the best overall performance, outperforming both the ResNet and DenseNet families as well as the larger ConvNeXt variants. Compared with the best-performing ResNet50 configuration, it reduces MAE by 29.04\% and RMSE by 22.52\%. Similarly, relative to the best DenseNet161 configuration, it reduces MAE by 46.10\% and RMSE by 39.56\%. Notably, although all previous experiments were conducted using ConvNeXt-B, the smaller ConvNeXt-T achieves comparable performance with fewer parameters (28M vs 88M) and slightly better results. These findings indicate that ConvNeXt-T is particularly well suited for this task and that increasing model capacity alone does not necessarily improve performance.

\paragraph{Hyperparameter Analysis.} 

Finally, we evaluate the influence of the remaining hyperparameters, namely the dropout probability in the range $[0.1, 0.5]$ and the weight decay in the range $[1 \times 10^{-2}, 1 \times 10^{-4}]$. As shown in Table~\ref{tab:ablation}, optimizing these parameters for the selected architecture and training strategy further improves performance, reducing MAE by 17.23\% and RMSE by 16.67\%.





\subsection{Annotation Reliability}

Before evaluating the memorization detection methods, we assessed the reliability of our manually annotated test set $\mathcal{P}_{\mathrm{test}}$. The obtained Fleiss' $\kappa$ score of 0.99 indicates almost perfect agreement among the three expert observers, confirming that the annotation protocol produces highly reproducible labels. The few disagreements observed were strictly confined to neighboring categories, suggesting uncertainty only in anatomically borderline cases rather than inconsistencies in the labeling criteria.


\begin{table*}[t]
\centering
\setlength{\tabcolsep}{6pt}
\def\arraystretch{1.2}
\resizebox{\textwidth}{!} {
    \begin{tabular}{cc|cc|cc|cc|ccc}

    \toprule
    \multirow{2}{*}{} & \multirow{2}{*}{\textbf{Method}}
    & \multicolumn{2}{c|}{\textcolor{teal}{\textbf{Unrelated}}}
    & \multicolumn{2}{c|}{\textcolor{orange}{\textbf{Near-Duplicate}}}
    & \multicolumn{2}{c|}{\textcolor{red}{\textbf{Exact Duplicate}}}
    & \multirow{2}{*}{\textbf{Macro F1}}
    & \multirow{2}{*}{\textbf{TPR@5\%FPR}}
    & \multirow{2}{*}{\textbf{Silhouette}}\\
    & & \textbf{Precision} & \textbf{Recall}
    & \textbf{Precision} & \textbf{Recall}
    & \textbf{Precision} & \textbf{Recall}
    & & & \\
    
    \midrule
    \multirow{7}{*}{\rotatebox[origin=c]{90}{\parbox{2.7cm}{\centering \textbf{(A) Perfectly Registered Pairs}}}}
    & SemDeDup & 71.82\% & \best{99.66\%} & 61.79\% & 2.57\% & 0.00\% & 0.00\% & 29.48\% & 24.93\% & -0.04 \\
    & UMS ($\ell_2$) & 95.64\% & 96.94\% & 86.69\% & 75.79\% & 56.59\% & \second{83.63\%} & 81.55\% & 90.64\% & 0.23 \\
    & UMS (SSCD) & 94.56\% & 90.08\% & 64.63\% & 67.99\% & 33.50\% & 49.61\% & 66.18\% & 54.02\% & 0.08 \\
    & Dar et al. & 71.49\% & \second{99.39\%} & N/A & N/A & 17.50\% & 1.81\% & 28.87\% & 23.11\% & 0.13 \\
    & SSIM & \best{99.66\%} & \best{99.66\%} & \best{95.24\%} & \best{98.31\%} & \best{94.62\%} & 77.66\% & \best{93.91\%} & \best{100.00\%} & \best{0.46} \\
    & DeepSSIM & 95.17\% & 94.62\% & 75.91\% & 52.57\% & 32.53\% & \best{91.16\%} & 68.33\% & 71.94\% & 0.31 \\
    & DeepSSIM++ (Proposed) & \second{96.28\%} & 97.61\% & \second{87.06\%} & \second{84.86\%} & \second{74.79\%} & 68.58\% & \second{84.81\%} & \second{94.29\%} & \second{0.39} \\
    \midrule
    
    \multirow{7}{*}{\rotatebox[origin=c]{90}{\parbox{2.7cm}{\centering \textbf{(B) Randomly Perturbed Pairs}}}}
    & SemDeDup & 74.13\% & 94.48\% & 47.40\% & 17.89\% & 0.00\% & 0.00\% & 36.35\% & \second{65.89\%} & -0.06 \\
    & UMS ($\ell_2$) & 71.95\% & \second{99.79\%} & 76.04\% & 3.41\% & \second{66.66\%} & 1.55\% & 31.06\% & 10.12\% & 0.00 \\
    & UMS (SSCD) & 76.32\% & 98.49\% & 69.03\% & 21.77\% & 39.39\% & 3.37\% & 41.77\% & 27.53\% & 0.02 \\
    & Dar et al. & 71.42\% & 98.72\% & N/A & N/A & 19.23\% & 1.29\% & 28.58\% & 14.54\% & 0.10 \\
    & SSIM & 71.78\% & \best{99.98\%} & \second{83.22\%} & 2.61\% & 0.00\% & 0.00\% & 29.54\% & 11.68\% & 0.06 \\
    & DeepSSIM & \second{95.61\%} & 92.63\% & 69.97\% & \second{55.42\%} & 31.31\% & \best{83.37\%} & \second{67.16\%} & 59.48\% & \second{0.27} \\
    & DeepSSIM++ (Proposed) & \best{96.35\%} & 96.69\% & \best{83.58\%} & \best{84.82\%} & \best{69.88\%} & \second{58.45\%} & \best{81.39\%} & \best{91.17\%} & \best{0.36} \\
    \bottomrule

    \end{tabular}
}
\caption{ \label{tab:brainmri} Quantitative performance evaluation on the annotated test set using precision and recall for each class, macro F1 score, TPR@5\%FPR, and Silhouette score. Results are reported for both perfectly registered pairs and randomly perturbed pairs. The best results for each metric are highlighted with a gray background and bold font, whereas the second-best results are highlighted with a gray background only. }
\end{table*}


\subsection{Quantitative Comparison against Baselines}

We perform a quantitative evaluation comparing our approach against baseline methods by assigning each image pair to one of three predefined similarity categories: \textit{Unrelated}, \textit{Near-Duplicate}, or \textit{Exact Duplicate}. These experiments are conducted on the annotated dataset $\mathcal{P}_{\mathrm{test}}$ described in Section~\ref{sec:pairs_dataset}. We report precision, recall, macro F1 score, and True Positive Rate at 5\% False Positive Rate (TPR@5\%FPR) for each evaluated method. Since TPR@5\%FPR is a binary metric, we adopt a one-vs-rest evaluation scheme, with \textit{Exact Duplicate} as the positive class. Additionally, we compute the Silhouette score to evaluate class separability in the learned feature space, providing a threshold-independent assessment of representation quality. All experiments are conducted under two settings: (A) an ideal setting with perfectly registered image pairs, and (B) a realistic challenging setting where image pairs undergo spatial and intensity perturbations. In the latter, images are subjected to random, anatomy-preserving transformations, including horizontal and vertical flips, subtle rotations, and contrast variations. Statistical significance is assessed using McNemar’s test on paired classification outcomes, with all improvements of DeepSSIM++ over competing approaches reaching significance ($p < 0.05$).

\begin{figure*}[pos=t]
\centering
\includegraphics[width=0.85\linewidth]{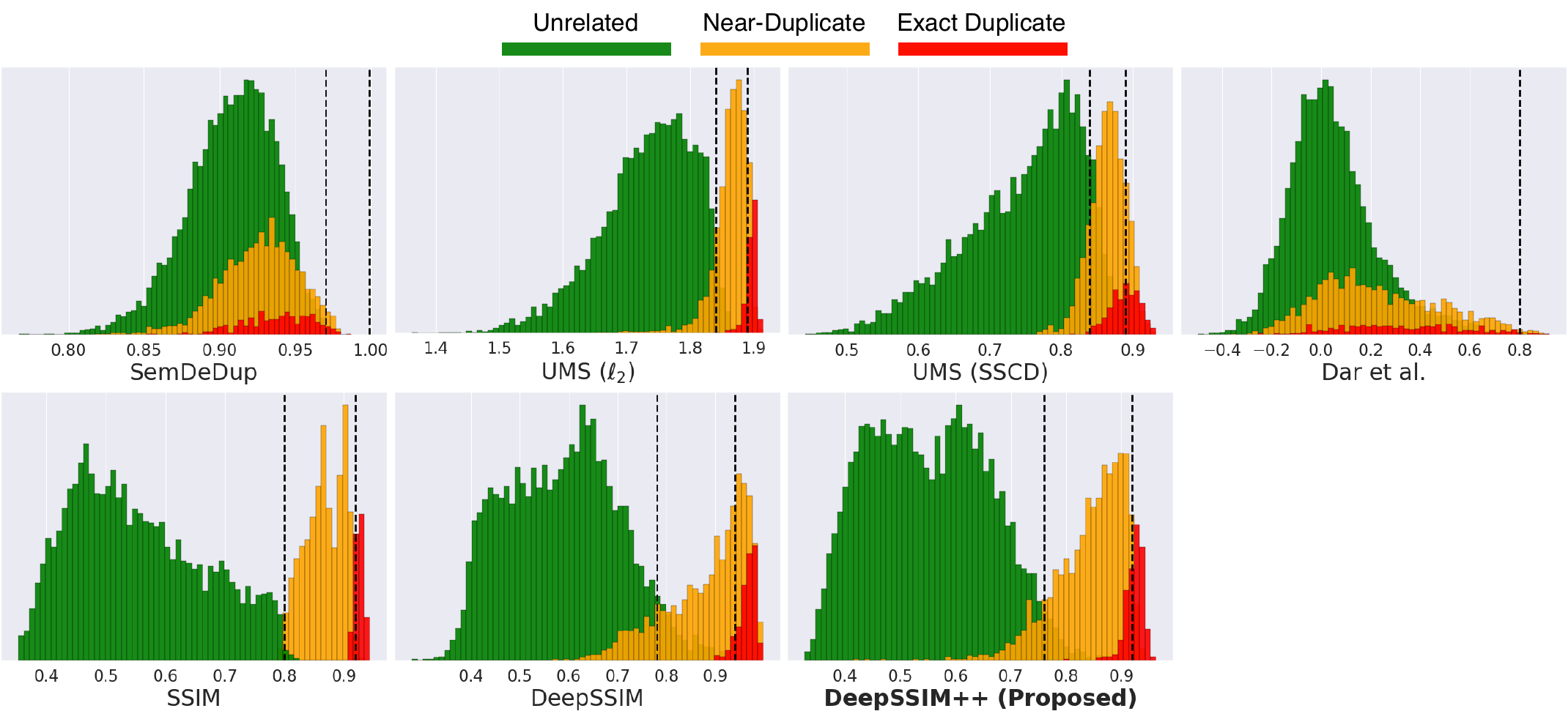}
\caption{ \label{fig:qualitative_brain} \textbf{Similarity Score Distributions Across Methods.} This figure illustrates the distribution of similarity scores predicted by each method under perfect registration, stratified by ground-truth category: \textit{Unrelated}, \textit{Near-Duplicate}, \textit{Exact Duplicate}. Dashed vertical lines indicate the decision threshold(s) used by each method. }
\end{figure*}

\subsubsection{Baselines}

The first baseline considered is the standard reference similarity measure SSIM. Although SSIM is not designed for memorization detection, its evaluation provides insight into the behavior of this traditional pixel-based metric under the two considered settings and serves as an approximate upper-bound reference in the ideal, alignment-dependent scenario rather than a direct competing baseline.

The second baseline is SemDeDup~\citep{abbas2023}, an approach originally designed for the semantic deduplication of large-scale training datasets, which we adapt to our evaluation protocol. Specifically, we map the ``perceptual duplicate'' category to our \textit{Exact Duplicate} class and the ``semantic duplicate'' category to our \textit{Near-Duplicate} class. To better capture biomedical image representations, we employ BioMedCLIP~\citep{zhang2024} as the underlying foundation model instead of the standard CLIP~\citep{radford2021} architecture originally used by the authors.

The next baseline is UMS~\citep{chen2026}, which categorizes image pairs into four similarity classes: \textit{different}, \textit{low}, \textit{medium}, and \textit{high}. UMS provides two variants based on different similarity measures: the normalized $\ell_2$ distance between embeddings for low-resolution inputs and SSCD embeddings for high-resolution inputs. To align UMS with our three-class setting, we merge the \textit{low} and \textit{medium} categories into a unified \textit{Near-Duplicate} class, while mapping \textit{high} similarity to \textit{Exact Duplicate} and \textit{different} to \textit{Unrelated}.

We also include the contrastive approach proposed by~\citet{dar2026}, which applies a single decision threshold based on the 95th percentile of the training--validation similarity distribution. As a result, this baseline performs binary classification between \textit{Exact Duplicate} and \textit{Unrelated} pairs, without explicitly identifying \textit{Near-Duplicate} cases. Using the official implementation, we retrained this method on our dataset following the authors' training protocol as closely as possible. Our final baseline is the original DeepSSIM model presented in~\citet{scardace2026}. 

To ensure a fair comparison across methods, we standardize the threshold selection procedure for approaches that rely on predefined decision thresholds, namely UMS and the original DeepSSIM. For each of these methods, we perform an independent grid search on the validation set to identify the optimal thresholds. SemDeDup and the contrastive method of~\citet{dar2026} are instead evaluated using their original decision procedures, as their classification mechanisms are integral to the methods themselves. Results obtained using the original thresholds reported for UMS and DeepSSIM are additionally provided in Appendix~\ref{appendix:benchmark} to quantify the effect of threshold adaptation.

\subsubsection{Performance on Registered Pairs}

As shown in Table~\ref{tab:brainmri} (A), under ideal registration conditions, DeepSSIM++ achieves a Silhouette score of $0.39$, outperforming all external memorization detection baselines and closely approaching the SSIM reference ($0.46$). Compared with the strongest external competitor, UMS ($\ell_2$), DeepSSIM++ achieves a substantial improvement in Silhouette score ($0.39$ vs $0.23$), indicating a more discriminative similarity space. This improved class separation is reflected in the classification results: DeepSSIM++ achieves an overall macro F1 score of $84.81\%$, outperforming the strongest external baseline, UMS ($\ell_2$), which reaches $81.55\%$. Likewise, DeepSSIM++ achieves a TPR@5\%FPR of $94.29\%$, compared with $90.64\%$ for UMS ($\ell_2$). The near-perfect performance of SSIM in this setting is expected, as images are perfectly aligned, confirming that pixel-wise similarity remains highly effective under ideal registration conditions. Relative to the original DeepSSIM formulation~\citep{scardace2026}, DeepSSIM++ substantially improves macro F1 from $68.33\%$ to $84.81\%$ and Silhouette score from $0.31$ to $0.39$. Although DeepSSIM++ exhibits lower \textit{Exact Duplicate} recall than the baseline ($68.58\%$ vs $91.16\%$), it substantially improves precision ($74.79\%$ vs $32.53\%$). By adopting a more conservative decision boundary, the model reclassifies borderline cases as \textit{Near-Duplicate}, yielding markedly higher recall in this category ($84.86\%$ vs $52.57\%$). This trade-off is well suited to auditing, as it reduces false positives while still surfacing suspicious samples for manual review.

\begin{figure*}[t]
\centering
\includegraphics[width=\textwidth]{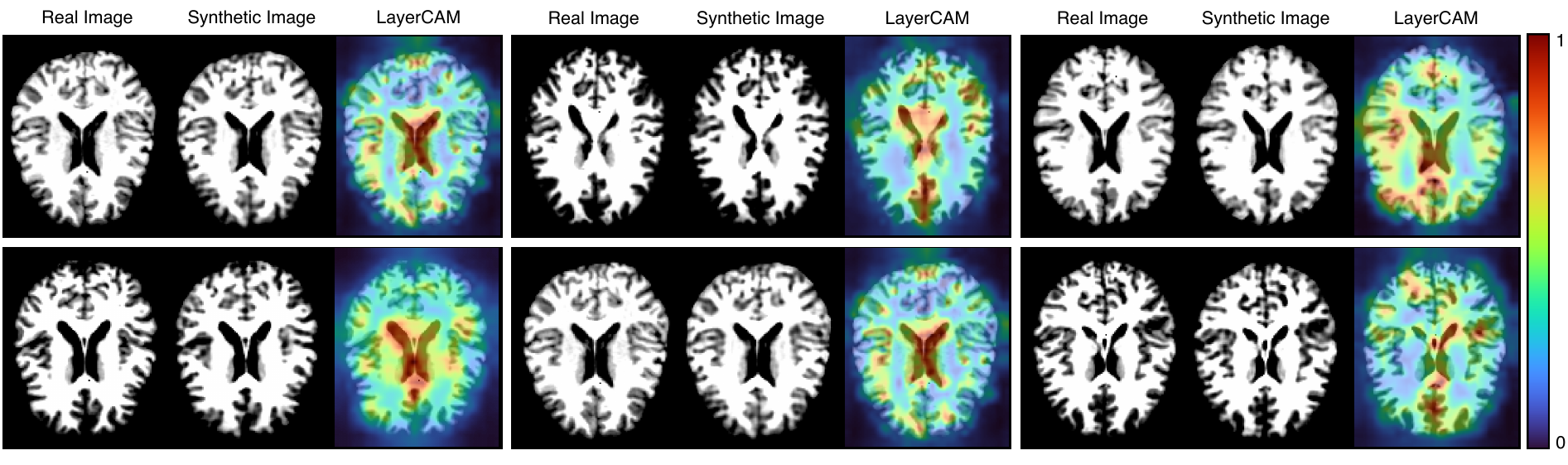}
\caption{\label{fig:layercam} \textbf{Qualitative Explainability Analysis via LayerCAM.} This figure illustrates six representative examples, comprising three \textit{Exact Duplicate} examples (top row) and three \textit{Near-Duplicate} examples (bottom row). For each pair, we show the real image, the corresponding synthetic image, and the associated LayerCAM activation map, which highlights the anatomical regions that most strongly drive the similarity score prediction and not the regions of agreement or discrepancy between the two images. Across both similarity categories, the network consistently attends to anatomically relevant structures, including the lateral ventricles, periventricular white matter, and cortical gyri, with negligible activation over background regions. }
\end{figure*}

\subsubsection{Performance on Perturbed Pairs}

As shown in Table~\ref{tab:brainmri} (B), DeepSSIM++ demonstrates strong robustness to spatial and contrast variations, achieving the highest overall Silhouette score ($0.36$), substantially exceeding the SSIM reference ($0.06$) by a factor of $6\times$. This result highlights the sensitivity of direct image-space similarity measures, such as SSIM and $\ell_2$ distance, to even small spatial perturbations, and the advantage of learned similarity representations that are less dependent on exact pixel alignment. DeepSSIM++ achieves an overall macro F1 score of $81.39\%$, outperforming the strongest external memorization detection baseline, UMS (SSCD), which reaches $41.77\%$, corresponding to a relative improvement of $95\%$. Likewise, DeepSSIM++ achieves a TPR@5\%FPR of $91.17\%$, compared with $65.89\%$ for SemDeDup, corresponding to a relative improvement of $38\%$. Interestingly, SemDeDup’s macro F1 increases under perturbations, rising from $29.48\%$ to $36.35\%$, primarily due to higher \textit{Near-Duplicate} recall. We hypothesize that this improvement does not reflect genuine robustness to structural changes. Instead, it may result from a distributional shift in which geometric and intensity perturbations move borderline embeddings across the decision threshold, causing previously missed samples to be classified as \textit{Near-Duplicate}. Notably, UMS ($\ell_2$), which represents one of the strongest baselines under ideal registration, substantially degrades under perturbations, reaching a Silhouette score of $0.00$, a macro F1 score of $31.06\%$, and a TPR@5\%FPR of only $10.12\%$. This degradation further highlights the limitations of direct image-space similarity measures under realistic spatial misalignment. In contrast, the original DeepSSIM formulation~\citep{scardace2026} remains substantially more robust than most competing approaches, while DeepSSIM++ further improves upon it, increasing macro F1 from $67.16\%$ to $81.39\%$ and Silhouette score from $0.27$ to $0.36$.

These results demonstrate that the proposed refinements improve the ability of DeepSSIM++ to learn alignment-tolerant similarity representations suitable for realistic, unregistered medical imaging scenarios.


\subsection{Qualitative Analysis}

Figure~\ref{fig:qualitative_brain} illustrates the similarity score distributions produced by each method across different classes. 
The resulting distributions indicate that baseline methods provide more limited separation between similarity categories, whereas DeepSSIM++ produces better-separated score distributions, consistent with the Silhouette scores reported in Table~\ref{tab:brainmri}.

\begin{figure}[pos=t]
\centering
\includegraphics[width=0.8\linewidth]{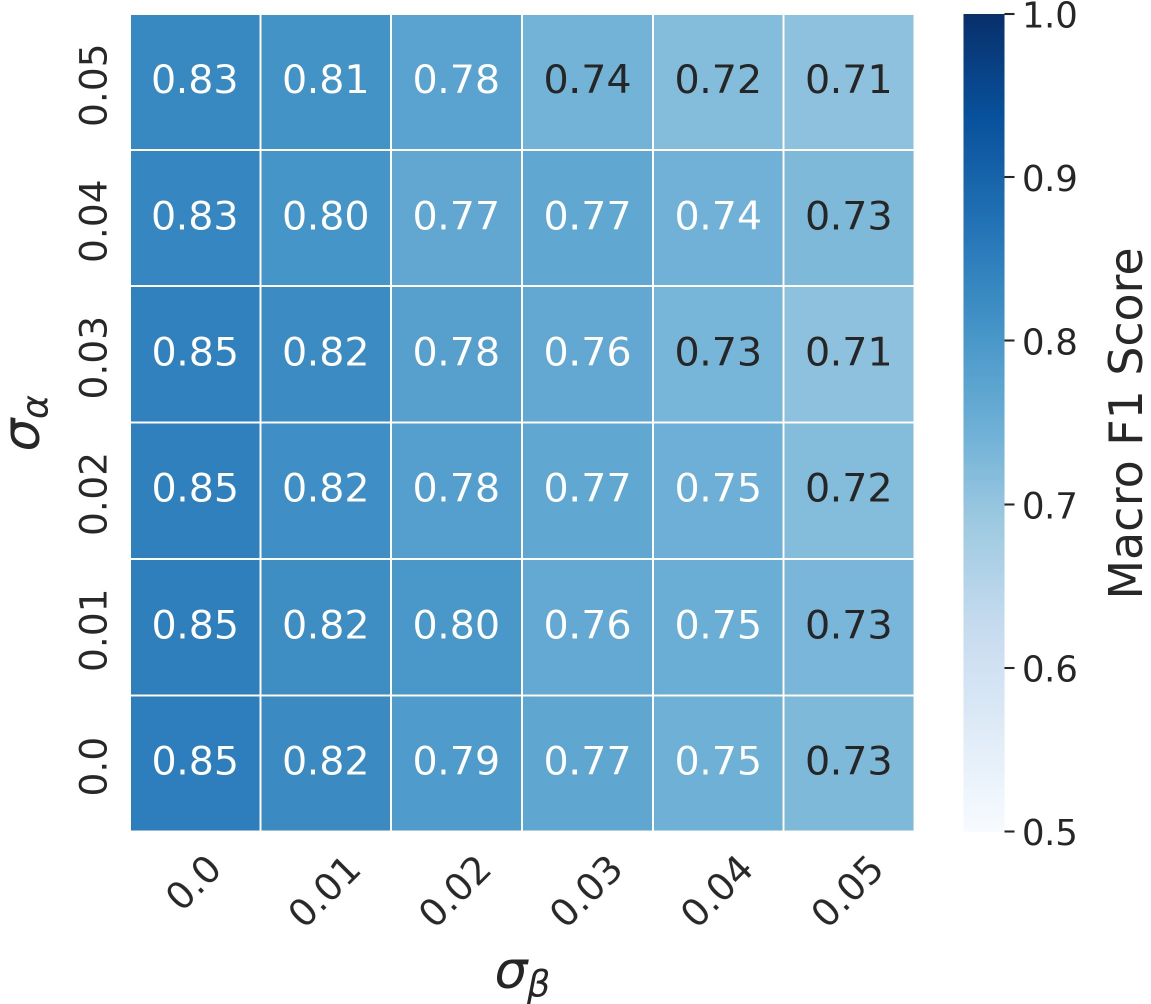}
\caption{\label{fig:thresholds}\textbf{Sensitivity of the Proposed Thresholds.} This figure illustrates the variations of the macro F1 score as a function of Gaussian noise applied to thresholds $\alpha$ and $\beta$. The $x$- and $y$-axes represent increasing noise levels, corresponding to $\sigma_\beta$ and $\sigma_\alpha$, respectively.}
\end{figure}

\subsubsection{Explainability Analysis}

To investigate whether DeepSSIM++ relies on meaningful visual regions when estimating image similarity, we perform an interpretability analysis using LayerCAM~\citep{jiang2021}. In particular, given a real–synthetic image pair, LayerCAM is obtained by backpropagating the predicted similarity score $s_\theta$. The resulting multi-scale activation maps are then aggregated and min–max normalized to $[0,1]$ for visualization. The activation map highlights regions considered by the network when producing the similarity score.

Figure~\ref{fig:layercam} shows representative explanations for six image pairs, covering both \textit{Exact Duplicate} (top row) and \textit{Near-Duplicate} (bottom row) cases. Across both categories, DeepSSIM++ consistently attends to anatomically relevant structures, most notably the lateral ventricles, periventricular white matter, and cortical gyri, with negligible activation over background regions. This pattern indicates that the network relies on anatomically meaningful evidence to estimate similarity regardless of the underlying degree of correspondence between the paired images. These results support the use of DeepSSIM++ as an interpretable auditing tool rather than a black-box model.


\subsection{Threshold Sensitivity Analysis}

To evaluate the robustness of the proposed thresholds $\alpha$ and $\beta$ in the DeepSSIM++ pipeline, we analyze their sensitivity to Gaussian perturbations using the annotated brain MRI test set $\mathcal{P}_{\mathrm{test}}$. Formally, let $\alpha=0.76$ and $\beta=0.92$ be the optimal threshold values. We define the perturbed thresholds $\tilde{\alpha}$ and $\tilde{\beta}$ by injecting independent additive noise:

\begin{equation}
    \tilde{\alpha} = \alpha + \epsilon_\alpha, \quad \tilde{\beta} = \beta + \epsilon_\beta
\end{equation}

where $\epsilon_\alpha \sim \mathcal{N}(0, \sigma_\alpha^2)$ and $\epsilon_\beta \sim \mathcal{N}(0, \sigma_\beta^2)$ are zero-mean Gaussian variables parameterized by their respective standard deviations $\sigma_\alpha$ and $\sigma_\beta$. Perturbations violating $\tilde{\alpha}<\tilde{\beta}$ or $\tilde{\alpha},\tilde{\beta} \in [0,1]$ are discarded.

We evaluate the impact of increasing noise levels, with $\sigma_\alpha$ and $\sigma_\beta$ ranging from 0.00 to 0.05 in increments of 0.01, by measuring the corresponding variation in macro F1 score. As shown in Figure~\ref{fig:thresholds}, the F1 score decreases gradually as perturbation magnitude increases, from 0.85 in the noise-free setting ($\sigma_\alpha = \sigma_\beta = 0$) to 0.71 under the largest joint perturbation ($\sigma_\alpha = \sigma_\beta = 0.05$), with no abrupt collapse observed anywhere in the grid. Notably, even under the largest perturbations, the performance consistently remains above all embedding-based baselines reported in Table~\ref{tab:brainmri}.


\subsection{Computational Efficiency}

Beyond its superior predictive performance, DeepSSIM++ offers a substantial computational advantage over traditional SSIM. By encoding images into compact vector embeddings, it enables the use of highly optimized nearest-neighbor search frameworks, such as FAISS~\citep{johnson2019}, for large-scale retrieval. Furthermore, cosine similarities between entire sets of real and synthetic images can be computed efficiently through a single matrix multiplication of their $\ell_2$-normalized embedding matrices, fully leveraging GPU-accelerated linear algebra routines.

\begin{figure}[pos=t]
\centering
\includegraphics[width=0.8\linewidth]{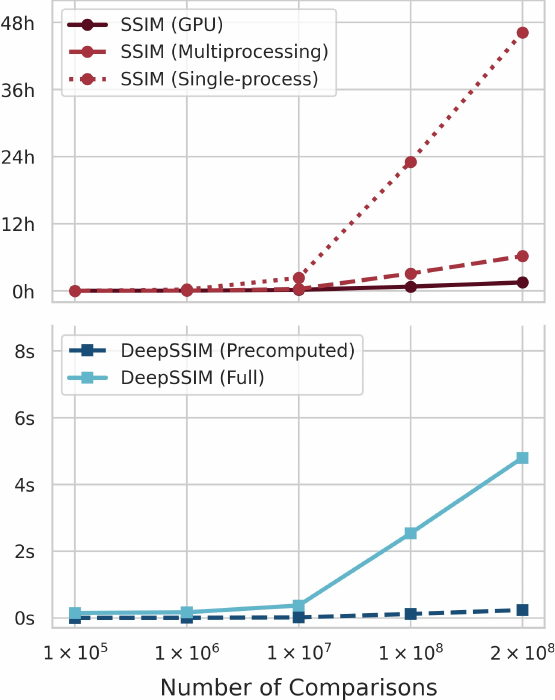}
\caption{ \label{fig:runtime} \textbf{Computational Runtime Comparison Between SSIM and DeepSSIM++.} This figure illustrates the elapsed time required to compute an increasing number of pairwise comparisons, shown on the $x$-axis. The top panel reports the runtime of three SSIM implementations: single-process (light blue), multiprocessing (dark blue, dashed), and GPU-accelerated (light green, dotted), measured in hours. The bottom panel reports DeepSSIM++ runtime under two configurations: full computation, including embedding extraction (light red), and pre-computed embeddings, requiring only similarity computation (dark red, dashed), measured in seconds. }
\end{figure}

In particular, Figure~\ref{fig:runtime} compares the execution time of SSIM and DeepSSIM++ as the number of image pairs increases. For SSIM, we report three implementations: the standard single-process version, a multiprocessing implementation, and a GPU-accelerated implementation. At small scales ($1 \times 10^5$ comparisons), all methods complete within seconds, and runtime differences remain limited. As the number of comparisons increases, however, the gap progressively widens, reflecting the fundamentally different computational complexity of pixel-wise similarity evaluation and embedding-based similarity computation. At the largest evaluated scale ($\sim2\times10^8$ image pairs), the standard single-process implementation of SSIM requires nearly 48 hours, while multiprocessing reduces the runtime to approximately 7 hours. Even the GPU implementation requires about 90 minutes to complete all pairwise comparisons. In contrast, DeepSSIM++ completes the entire pipeline, including embedding extraction and similarity computation, in approximately 5 seconds, or in less than 1 second when embeddings are pre-computed. Compared with the GPU implementation of SSIM, this corresponds to approximately three orders of magnitude faster, while the speedup exceeds four orders of magnitude relative to the single-process implementation. Since embeddings are computed only once, subsequent similarity queries require only a matrix multiplication in the embedding space. This makes DeepSSIM++ readily scalable to datasets containing hundreds of millions of image comparisons, as demonstrated in this work.

\section{Discussion and Conclusion} \label{sec:discussion}

In this work, we introduced DeepSSIM++, a self-supervised similarity metric for detecting memorization in medical generative models. Our results demonstrate that DeepSSIM++ achieves performance comparable to SSIM under ideal registration conditions, while substantially outperforming SSIM and all evaluated learning-based baselines when image pairs undergo realistic spatial and intensity perturbations. This robustness is particularly relevant for practical memorization auditing, where perfect pixel-level alignment between real and synthetic images cannot generally be assumed. Beyond detection performance, DeepSSIM++ provides a computational advantage of several orders of magnitude over pixel-space SSIM at scale, enabling memorization analysis on datasets involving hundreds of millions of image pairs, for which exhaustive pixel-wise evaluation becomes impractical. The explainability analysis further indicates that DeepSSIM++ predictions are grounded in anatomically meaningful structures rather than spurious visual patterns, supporting its use as an interpretable auditing tool rather than a black-box classifier.

Despite these strengths, we acknowledge that the decision thresholds $\alpha$ and $\beta$ are calibrated on our brain MRI dataset, as the mapping from continuous similarity scores to memorization categories is data-distribution dependent. Adaptive thresholding strategies based on unsupervised density estimation, such as Gaussian Mixture Models or multi-level Otsu thresholding, provide potential alternatives but rely on the assumption that memorization categories correspond to statistically separable modes in the score distribution. This assumption cannot be guaranteed, as in a continuous similarity space, semantic categories may overlap, collapse into a single density mode, or exhibit multiple modes within the same category depending on dataset composition. Model selection criteria such as the Bayesian Information Criterion can estimate the number of statistical components but cannot determine their correspondence with meaningful memorization classes. Consequently, while the similarity metric is expected to generalize across domains, the decision thresholds may require recalibration to adapt to the specific score distribution of any new dataset.

Beyond memorization detection, the effectiveness and scalability of DeepSSIM++ enable additional practical applications in privacy-preserving generation. For example, before the public release of synthetic datasets, generated samples could be automatically screened to identify and remove those classified as near- or exact-duplicates, reducing the risk of unintended training data disclosure. More interestingly, DeepSSIM++ could be integrated directly into the generation pipeline: embeddings of training images could be pre-computed, and during inference, each newly generated sample could be mapped into the embedding space and compared against the training set to identify its nearest neighbors. Samples exhibiting near- or exact-duplicate similarity could then be discarded. Lastly, our approach could be used by experts in medical imaging to evaluate how much their models memorize training data.

Future research will aim to integrate DeepSSIM++ directly into generative model training, moving beyond post-hoc detection toward generation procedures that explicitly control similarity to the training set. In this setting, DeepSSIM++ could serve as a regularization term to penalize the reproduction of training samples, mitigating memorization while preserving data utility and ensuring the generation of diverse and clinically meaningful outputs.


\appendix

\section{Results Using Original Threshold Values} \label{appendix:benchmark}

To complement the standardized evaluation in the main text, we evaluate UMS and the original DeepSSIM using the decision thresholds reported in their respective publications. Unlike the main-text results, where thresholds were selected via grid search on the validation set for fair comparison, these results assess the out-of-the-box applicability of the original methods to the categories defined in Section~\ref{sec:labels}. As shown in Table~\ref{tab:brainmri_appendix}, these thresholds substantially degrade performance: UMS ($\ell_2$ and SSCD) achieves macro F1 below $3.3\%$, while the original DeepSSIM drops from $68.33\%$ to $41.32\%$ on registered pairs and from $67.16\%$ to $40.48\%$ under perturbations, due to overly aggressive \textit{Exact Duplicate} classification. Importantly, the advantage of DeepSSIM++ is not attributable to threshold calibration alone. Metrics independent of the decision threshold favor DeepSSIM++, with higher Silhouette score ($0.39$ vs $0.23$ for UMS $\ell_2$) and TPR@5\%FPR ($94.29\%$ vs $90.65\%$), confirming a more discriminative similarity space.

\begin{table*}[t]
\centering
\setlength{\tabcolsep}{6pt}
\def\arraystretch{1.2}
\resizebox{\textwidth}{!} {
    \begin{tabular}{cc|cc|cc|cc|ccc}

    \toprule
    \multirow{2}{*}{} & \multirow{2}{*}{\textbf{Method}}
    & \multicolumn{2}{c|}{\textcolor{teal}{\textbf{Unrelated}}}
    & \multicolumn{2}{c|}{\textcolor{orange}{\textbf{Near-Duplicate}}}
    & \multicolumn{2}{c|}{\textcolor{red}{\textbf{Exact Duplicate}}}
    & \multirow{2}{*}{\textbf{Macro F1}}
    & \multirow{2}{*}{\textbf{TPR@5\%FPR}}
    & \multirow{2}{*}{\textbf{Silhouette}}\\
    & & \textbf{Precision} & \textbf{Recall}
    & \textbf{Precision} & \textbf{Recall}
    & \textbf{Precision} & \textbf{Recall}
    & & & \\
    
    \midrule
    \multirow{7}{*}{\rotatebox[origin=c]{90}{\parbox{2.7cm}{\centering \textbf{(A) Perfectly Registered Pairs}}}}
    & SemDeDup & 71.82\% & 99.66\% & 61.79\% & 2.57\% & 0.00\% & 0.00\% & 29.48\% & 24.93\% & -0.04 \\
    & UMS ($\ell_2$) & 0.00\% & 0.00\% & 0.00\% & 0.00\% & 4.39\% & 100.00\% & 2.80\% & 90.64\% & 0.23 \\
    & UMS (SSCD) & 0.00\% & 0.00\% & 0.00\% & 0.00\% & 4.60\% & 100.00\% & 2.93\% & 54.02\% & 0.08 \\
    & Dar et al. & 71.49\% & 99.39\% & N/A & N/A & 17.50\% & 1.81\% & 28.87\% & 23.11\% & 0.13 \\
    & SSIM & 99.66\% & 99.66\% & 95.24\% & 98.31\% & 94.62\% & 77.66\% & 93.91\% & 100.00\% & 0.46 \\
    & DeepSSIM & 99.93\% & 52.48\% & 18.30\% & 29.96\% & 19.38\% & 100.00\% & 41.32\% & 71.94\% & 0.31 \\
    & DeepSSIM++ (Proposed) & 96.28\% & 97.61\% & 87.06\% & 84.86\% & 74.79\% & 68.58\% & 84.81\% & 94.29\% & 0.39 \\
    \midrule
    
    \multirow{7}{*}{\rotatebox[origin=c]{90}{\parbox{2.7cm}{\centering \textbf{(B) Randomly Perturbed Pairs}}}}
    & SemDeDup & 74.13\% & 94.48\% & 47.40\% & 17.89\% & 0.00\% & 0.00\% & 36.35\% & 65.89\% & -0.06 \\
    & UMS ($\ell_2$) & 0.00\% & 0.00\% & 0.00\% & 0.00\% & 4.39\% & 100.00\% & 2.80\% & 10.12\% & 0.00 \\
    & UMS (SSCD) & 100.00\% & 0.15\% & 0.10\% & 0.04\% & 4.93\% & 100.00\% & 3.26\% & 27.53\% & 0.02 \\
    & Dar et al. & 71.42\% & 98.72\% & N/A & N/A & 19.23\% & 1.29\% & 28.58\% & 14.54\% & 0.10 \\
    & SSIM & 71.78\% & 99.98\% & 83.22\% & 2.61\% & 0.00\% & 0.00\% & 29.54\% & 11.68\% & 0.06 \\
    & DeepSSIM & 99.93\% & 50.49\% & 17.80\% & 29.95\% & 19.08\% & 100.00\% & 40.48\% & 59.48\% & 0.27 \\
    & DeepSSIM++ (Proposed) & 96.35\% & 96.69\% & 83.58\% & 84.82\% & 69.88\% & 58.45\% & 81.39\% & 91.17\% & 0.36 \\
    \bottomrule
    
    \end{tabular}
}
\caption{ \label{tab:brainmri_appendix} Quantitative performance evaluation on the annotated test set using precision and recall for each class, macro F1 score, TPR@5\%FPR, and Silhouette score. Results are reported for both perfectly registered pairs and randomly perturbed pairs, using the original decision thresholds reported for UMS and DeepSSIM. The best results for each metric are highlighted with a gray background and bold font, whereas the second-best results are highlighted with a gray background only. }
\end{table*}


\printcredits
\bibliographystyle{cas-model2-names}
\bibliography{cas-refs}

@article{ddpm2020,
    author={Ho, Jonathan and Jain, Ajay and Abbeel, Pieter},
    title={Denoising diffusion probabilistic models},
    journal={Advances in neural information processing systems},
    volume={33},
    pages={6840--6851},
    year={2020}
}

@inproceedings{ldm2022,
    author={Rombach, Robin and Blattmann, Andreas and Lorenz, Dominik and Esser, Patrick and Ommer, Björn}, 
    title={High-Resolution Image Synthesis with Latent Diffusion Models}, 
    booktitle={2022 IEEE/CVF Conference on Computer Vision and Pattern Recognition (CVPR)},
    year={2022},
    pages={10674-10685},
    doi={10.1109/CVPR52688.2022.01042}
}

@inproceedings{lipman2023,
    title={Flow Matching for Generative Modeling},
    author={Yaron Lipman and Ricky T. Q. Chen and Heli Ben-Hamu and Maximilian Nickel and Matthew Le},
    booktitle={The Eleventh International Conference on Learning Representations },
    year={2023},
    url={https://openreview.net/forum?id=PqvMRDCJT9t}
}

@inproceedings{gan2014,
    author={Goodfellow, Ian and Pouget-Abadie, Jean and Mirza, Mehdi and Xu, Bing and Warde-Farley, David and Ozair, Sherjil and others},
    title={Generative adversarial nets},
    booktitle={Advances in Neural Information Processing Systems (NeurIPS)},
    volume={27},
    year={2014}
}

@inproceedings{vae2014,
    author={Diederik P. Kingma and Max Welling},
    title={Auto-Encoding Variational Bayes},
    booktitle={2nd International Conference on Learning Representations, {ICLR} 2014, Banff, AB, Canada, April 14-16, 2014, Conference Track Proceedings},
    year={2014}
}

@article{ahishakiye2021,
    author = {Emmanuel Ahishakiye and Martin Bastiaan {Van Gijzen} and Julius Tumwiine and Ruth Wario and Johnes Obungoloch},
    title = {A survey on deep learning in medical image reconstruction},
    journal = {Intelligent Medicine},
    volume = {1},
    number = {3},
    pages = {118-127},
    year = {2021},
    doi = {https://doi.org/10.1016/j.imed.2021.03.003}
}

@inproceedings{sargood2026,
    author = {Alec Sargood and L. Puglisi and J. Cole and N. Oxtoby and D. Rav{\`\i}},
    title = {CoCoLIT: ControlNet-Conditioned Latent Image Translation for MRI to Amyloid PET Synthesis},
    booktitle = {Proceedings of the AAAI Conference on Artificial Intelligence},
    volume = {40},
    number = {1},
    year = {2026},
    url = {https://ojs.aaai.org/index.php/AAAI/article/view/37831}
}

@article{brlp2025,
    author = {Lemuel Puglisi and Daniel C. Alexander and Daniele Ravì},
    title = {Brain Latent Progression: Individual-based spatiotemporal disease progression on 3D Brain MRIs via latent diffusion},
    journal = {Medical Image Analysis},
    volume = {106},
    pages = {103734},
    year = {2025},
    doi = {https://doi.org/10.1016/j.media.2025.103734}
}

@inproceedings {carlini2023,
    author = {Nicolas Carlini and Jamie Hayes and Milad Nasr and Matthew Jagielski and Vikash Sehwag and Florian Tram{\`e}r and others},
    title = {Extracting Training Data from Diffusion Models},
    booktitle = {32nd USENIX Security Symposium (USENIX Security 23)},
    year = {2023},
    pages = {5253--5270},
    url = {https://www.usenix.org/conference/usenixsecurity23/presentation/carlini},
    publisher = {USENIX Association}
}

@inproceedings{arora2018,
    title={Do {GAN}s learn the distribution? Some Theory and Empirics},
    author={Sanjeev Arora and Andrej Risteski and Yi Zhang},
    booktitle={International Conference on Learning Representations},
    year={2018},
    url={https://openreview.net/forum?id=BJehNfW0-}
}

@article{gu2025,
    title={On Memorization in Diffusion Models},
    author={Xiangming Gu and Chao Du and Tianyu Pang and Chongxuan Li and Min Lin and Ye Wang},
    journal={Transactions on Machine Learning Research},
    year={2025},
    url={https://openreview.net/forum?id=D3DBqvSDbj}
}

@inproceedings{somepalli2023,
    author={Somepalli, Gowthami and Singla, Vasu and Goldblum, Micah and Geiping, Jonas and Goldstein, Tom},
    booktitle={2023 IEEE/CVF Conference on Computer Vision and Pattern Recognition (CVPR)}, 
    title={Diffusion Art or Digital Forgery? Investigating Data Replication in Diffusion Models}, 
    year={2023},
    volume={},
    number={},
    pages={6048-6058},
    doi={10.1109/CVPR52729.2023.00586}
}

@inproceedings{bonnaire2026,
    title={Why Diffusion Models Don{\textquoteright}t Memorize:  The Role of Implicit Dynamical Regularization in Training},
    author={Tony Bonnaire and Rapha{\"e}l Urfin and Giulio Biroli and Marc Mezard},
    booktitle={The Thirty-ninth Annual Conference on Neural Information Processing Systems},
    year={2026},
    url={https://openreview.net/forum?id=BSZqpqgqM0}
}

@inproceedings{yoon2023,
    title={Diffusion Probabilistic Models Generalize when They Fail to Memorize},
    author={TaeHo Yoon and Joo Young Choi and Sehyun Kwon and Ernest K. Ryu},
    booktitle={ICML 2023 Workshop on Structured Probabilistic Inference {\&} Generative Modeling},
    year={2023},
    url={https://openreview.net/forum?id=shciCbSk9h}
}

@article{gillotte2020,
    author = {Gillotte, Jessica},
    title = {Copyright Infringement in AI-Generated Artworks},
    journal = {UC Davis Law Review},
    volume = {53},
    number = {5},
    pages = {2655--2691},
    year = {2020},
    url = {https://ssrn.com/abstract=3657423}
}

@misc{eu2016gdpr,
    author = {{European Union}},
    title = {Regulation (EU) 2016/679 of the European Parliament and of the Council of 27 April 2016 on the protection of natural persons with regard to the processing of personal data and on the free movement of such data, and repealing Directive 95/46/EC (General Data Protection Regulation)},
    journal = {Official Journal of the European Union},
    volume = {L 119},
    pages = {1--88},
    year = {2016},
    url = {https://eur-lex.europa.eu/eli/reg/2016/679/oj}
}

@misc{eu2024aiact,
    author = {{European Union}},
    title = {Regulation (EU) 2024/1689 of the European Parliament and of the Council of 13 June 2024 laying down harmonised rules on artificial intelligence and amending Regulations (EC) No 300/2008, (EU) No 167/2013, (EU) No 168/2013, (EU) 2018/858, (EU) 2018/1139 and (EU) 2019/2144 and Directives 2014/90/EU, (EU) 2016/797 and (EU) 2020/1828 (Artificial Intelligence Act)},
    journal = {Official Journal of the European Union},
    volume = {L 2024/1689},
    year = {2024},
    url = {https://eur-lex.europa.eu/eli/reg/2024/1689/oj/eng}
}

@inproceedings{shokri2017,
    author={Shokri, Reza and Stronati, Marco and Song, Congzheng and Shmatikov, Vitaly},
    booktitle={2017 IEEE Symposium on Security and Privacy (SP)}, 
    title={Membership Inference Attacks Against Machine Learning Models}, 
    year={2017},
    volume={},
    number={},
    pages={3-18},
    doi={10.1109/SP.2017.41}
}

@inproceedings{carlini2022,
    author={Carlini, Nicholas and Chien, Steve and Nasr, Milad and Song, Shuang and Terzis, Andreas and Tramèr, Florian},
    booktitle={2022 IEEE Symposium on Security and Privacy (SP)}, 
    title={Membership Inference Attacks From First Principles}, 
    year={2022},
    pages={1897-1914},
    doi={10.1109/SP46214.2022.9833649}
}

@inproceedings{fredrikson2015,
    author = {Fredrikson, Matt and Jha, Somesh and Ristenpart, Thomas},
    title = {Model Inversion Attacks that Exploit Confidence Information and Basic Countermeasures},
    year = {2015},
    publisher = {Association for Computing Machinery},
    doi = {10.1145/2810103.2813677},
    booktitle = {Proceedings of the 22nd ACM SIGSAC Conference on Computer and Communications Security},
    pages = {1322–1333},
    numpages = {12}
}

@inproceedings{zhang2020,
    author={Zhang, Yuheng and Jia, Ruoxi and Pei, Hengzhi and Wang, Wenxiao and Li, Bo and Song, Dawn},
    booktitle={2020 IEEE/CVF Conference on Computer Vision and Pattern Recognition (CVPR)}, 
    title={The Secret Revealer: Generative Model-Inversion Attacks Against Deep Neural Networks}, 
    year={2020},
    volume={},
    number={},
    pages={250-258},
    doi={10.1109/CVPR42600.2020.00033}
}

@inproceedings{carlini2019,
    author = {Carlini, Nicholas and Liu, Chang and Erlingsson, \'{U}lfar and Kos, Jernej and Song, Dawn},
    title = {The secret sharer: evaluating and testing unintended memorization in neural networks},
    year = {2019},
    publisher = {USENIX Association},
    booktitle = {Proceedings of the 28th USENIX Conference on Security Symposium},
    pages = {267–284},
    numpages = {18}
}

@inproceedings{carlini2021,
    author = {Nicholas Carlini and Florian Tram{\`e}r and Eric Wallace and Matthew Jagielski and Ariel Herbert-Voss and Katherine Lee and others},
    title = {Extracting Training Data from Large Language Models},
    booktitle = {30th USENIX Security Symposium (USENIX Security 21)},
    year = {2021},
    pages = {2633--2650},
    url = {https://www.usenix.org/conference/usenixsecurity21/presentation/carlini-extracting},
    publisher = {USENIX Association}
}

@inproceedings{balle2021,
    title={Reconstructing Training Data with Informed Adversaries},
    author={Borja Balle and Giovanni Cherubin and Jamie Hayes},
    booktitle={NeurIPS 2021 Workshop Privacy in Machine Learning},
    year={2021},
    url={https://openreview.net/forum?id=Yi2DZTbnBl4}
}

@inproceedings{jahanian2022,
    title={Generative Models as a Data Source for Multiview Representation Learning},
    author={Ali Jahanian and Xavier Puig and Yonglong Tian and Phillip Isola},
    booktitle={International Conference on Learning Representations},
    year={2022},
    url={https://openreview.net/forum?id=qhAeZjs7dCL}
}

@article{agarwal2024,
	author = {Agarwal, S. and Wood, D. and Carpenter, R. and Wei, Y. and Modat, M. and Booth, T. C.},
    title = {Letter to the editor: what are the legal and ethical considerations of submitting radiology reports to ChatGPT?},
	doi = {10.1016/j.crad.2024.03.017},
	journal = {Clinical Radiology},
	number = {7},
	pages = {e979--e981},
	publisher = {Elsevier},
	volume = {79},
	year = {2024},
}

@article{wang2004,
    author = {Wang, Zhou and Bovik, Alan and Sheikh, H.R. and Simoncelli, Eero},
    title={Image quality assessment: from error visibility to structural similarity}, 
    journal={IEEE Transactions on Image Processing}, 
    year={2004},
    volume={13},
    number={4},
    pages={600-612},
    doi={10.1109/TIP.2003.819861}
}

@inproceedings{nasr2025,
    title={Scalable Extraction of Training Data from Aligned, Production Language Models},
    author={Milad Nasr and Javier Rando and Nicholas Carlini and Jonathan Hayase and Matthew Jagielski and A. Feder Cooper and others},
    booktitle={The Thirteenth International Conference on Learning Representations},
    year={2025},
    url={https://openreview.net/forum?id=vjel3nWP2a}
}

@article{dar2026,
    author = {Dar, Salman Ul Hassan and Seyfarth, Marvin and Ayx, Isabelle and Papavassiliu, Theano and Schoenberg, Stefan O. and Siepmann, Robert Malte and others},
    journal = {Nature Biomedical Engineering},
    number = {3},
    pages = {458--472},
    title = {Unconditional latent diffusion models memorize patient imaging data},
    volume = {10},
    year = {2026}
}

@inproceedings{dar2023,
    author = {Dar, Salman Ul Hassan and Ghanaat, Arman and Kahmann, Jannik and Ayx, Isabelle and Papavassiliu, Theano and Schoenberg, Stefan O. and others},
    title = {Investigating Data Memorization in 3D Latent Diffusion Models for Medical Image Synthesis},
    year = {2023},
    doi = {10.1007/978-3-031-53767-7_6},
    booktitle = {Deep Generative Models: Third MICCAI Workshop, DGM4MICCAI 2023, Held in Conjunction with MICCAI 2023, Vancouver, BC, Canada, October 8, 2023, Proceedings},
    pages = {56–65},
    numpages = {10}
}

@inproceedings{pizzi2022,
    author={Pizzi, Ed and Roy, Sreya Dutta and Ravindra, Sugosh Nagavara and Goyal, Priya and Douze, Matthijs},
    booktitle={2022 IEEE/CVF Conference on Computer Vision and Pattern Recognition (CVPR)}, 
    title={A Self-Supervised Descriptor for Image Copy Detection}, 
    year={2022},
    volume={},
    number={},
    pages={14512-14522},
    doi={10.1109/CVPR52688.2022.01413}
}

@inproceedings{breger2025,
    title="A Study on the Adequacy of Common IQA Measures for Medical Images",
    author="Breger, Anna and Karner, Clemens and Selby, Ian and Gr{\"o}hl, Janek and Dittmer, S{\"o}ren and Lilley, Edward and others",
    booktitle="Proceedings of 2024 International Conference on Medical Imaging and Computer-Aided Diagnosis (MICAD 2024)",
    year="2025",
    publisher="Springer Nature Singapore",
    pages="451--462"
}

@inproceedings{abadi2016,
    author = {Abadi, Martin and Chu, Andy and Goodfellow, Ian and McMahan, H. Brendan and Mironov, Ilya and Talwar, Kunal and others},
    title = {Deep Learning with Differential Privacy},
    year = {2016},
    url = {https://doi.org/10.1145/2976749.2978318},
    doi = {10.1145/2976749.2978318},
    booktitle = {Proceedings of the 2016 ACM SIGSAC Conference on Computer and Communications Security},
    pages = {308–318},
    numpages = {11}
}

@inproceedings{mcmahan2017,
    title = 	 {{Communication-Efficient Learning of Deep Networks from Decentralized Data}},
    author = 	 {McMahan, Brendan and Moore, Eider and Ramage, Daniel and Hampson, Seth and Arcas, Blaise Aguera y},
    booktitle = 	 {Proceedings of the 20th International Conference on Artificial Intelligence and Statistics},
    pages = 	 {1273--1282},
    year = 	 {2017},
    volume = 	 {54},
    series = 	 {Proceedings of Machine Learning Research},
    url = 	 {https://proceedings.mlr.press/v54/mcmahan17a.html}
}

@article{sheller2020,
	author = {Sheller, Micah J. and Edwards, Brandon and Reina, G. Anthony and Martin, Jason and Pati, Sarthak and Kotrotsou, Aikaterini and others},
	doi = {10.1038/s41598-020-69250-1},
	journal = {Scientific Reports},
	number = {1},
	pages = {12598},
	title = {Federated learning in medicine: facilitating multi-institutional collaborations without sharing patient data},
	volume = {10},
	year = {2020}
}

@article{ziller2021,
    author = {Ziller, Alexander and Usynin, Dmitrii and Braren, Rickmer and Makowski, Marcus and Rueckert, Daniel and Kaissis, Georgios},
    doi = {10.1038/s41598-021-93030-0},
    journal = {Scientific Reports},
    number = {1},
    pages = {13524},
    title = {Medical imaging deep learning with differential privacy},
    volume = {11},
    year = {2021}
}

@article{chauvin2020,
    author = {Laurent Chauvin and Kuldeep Kumar and Christian Wachinger and Marc Vangel and Jacques {de Guise} and Christian Desrosiers and others},
    title = {Neuroimage signature from salient keypoints is highly specific to individuals and shared by close relatives},
    journal = {NeuroImage},
    volume = {204},
    pages = {116208},
    year = {2020},
    doi = {https://doi.org/10.1016/j.neuroimage.2019.116208}
}

@inproceedings{scardace2026,
    author = { Scardace, Antonio and Puglisi, Lemuel and Guarnera, Francesco and Battiato, Sebastiano and Ravi, Daniele },
    booktitle = { 2026 IEEE/CVF Winter Conference on Applications of Computer Vision (WACV) },
    title = {{ A Novel Metric for Detecting Memorization in Generative Models for Brain MRI Synthesis }},
    year = {2026},
    pages = {3868-3877},
    doi = {10.1109/WACV61042.2026.00377},
    publisher = {IEEE Computer Society}
}

@article{chen2026,
    title={SIDE: Surrogate Conditional Data Extraction from Diffusion Models},
    author={Chen, Yunhao and Wang, Shujie and Zou, Difan and Ma, Xingjun},
    doi={10.1609/aaai.v40i1.36972},
    journal={Proceedings of the AAAI Conference on Artificial Intelligence},
    volume={40},
    pages={128–136},
    number={1},
    year={2026}
}

@misc{butterick2023,
    author = {Butterick, Matthew and {Joseph Saveri Law Firm}},
    title = {Stable Diffusion Litigation},
    year = {2023},
    url = {https://imagegeneratorlitigation.com/}
}

@misc{hipaaprivacyrule2003,
    author       = {{U.S. Department of Health and Human Services}},
    title        = {Standards for Privacy of Individually Identifiable Health Information (HIPAA Privacy Rule)},
    howpublished = {45 CFR Parts 160 and 164},
    year         = {2003}
}

@inproceedings{Seo2025,
    author = { Seo, Sumin and Lee, In Kyu and Kim, Hyun-Woo and Min, Jaesik and Jung, Chung-Hwan},
    title = { { Diffusion-Based User-Guided Data Augmentation for Coronary Stenosis Detection } },
    booktitle = {proceedings of Medical Image Computing and Computer Assisted Intervention -- MICCAI 2025},
    year = {2025},
    publisher = {Springer Nature Switzerland},
    volume = {LNCS 15967},
    month = {September},
    pages = {149 -- 169}
}

@inproceedings{feng2021,
    author={Feng, Qianli and Guo, Chenqi and Benitez-Quiroz, Fabian and Martinez, Aleix},
    booktitle={2021 IEEE/CVF International Conference on Computer Vision (ICCV)}, 
    title={When do GANs replicate? On the choice of dataset size}, 
    year={2021},
    volume={},
    number={},
    pages={6681-6690},
    doi={10.1109/ICCV48922.2021.00663}
}

@inproceedings{chen2024,
    author={Chen, Chen and Liu, Daochang and Xu, Chang},
    booktitle={2024 IEEE/CVF Conference on Computer Vision and Pattern Recognition (CVPR)}, 
    title={Towards Memorization-Free Diffusion Models}, 
    year={2024},
    volume={},
    number={},
    pages={8425-8434},
    doi={10.1109/CVPR52733.2024.00805}
}

@article{vincent2011,
    author={Vincent, Pascal},
    journal={Neural Computation}, 
    title={A Connection Between Score Matching and Denoising Autoencoders}, 
    year={2011},
    volume={23},
    number={7},
    pages={1661-1674},
    doi={10.1162/NECO_a_00142}
}

@article{gu2021,
    title={Domain-specific language model pretraining for biomedical natural language processing},
    author={Gu, Yu and Tinn, Robert and Cheng, Hao and Lucas, Michael and Usuyama, Naoto and Liu, Xiaodong and others},
    journal={ACM Transactions on Computing for Healthcare (HEALTH)},
    volume={3},
    number={1},
    pages={1--23},
    year={2021},
    publisher={ACM New York, NY}
}

@article{zhang2011fsim,
    author={Zhang, Lin and Zhang, Lei and Mou, Xuanqin and Zhang, David},
    journal={IEEE Transactions on Image Processing}, 
    title={FSIM: A Feature Similarity Index for Image Quality Assessment}, 
    year={2011},
    volume={20},
    number={8},
    pages={2378-2386},
    doi={10.1109/TIP.2011.2109730}
}

@article{bu2024,
    author = {Bu, Chenyang and Liu, Yuxin and Huang, Manzong and Shao, Jianxuan and Ji, Shengwei and Luo, Wenjian and others},
    title = {Layer-Wise Learning Rate Optimization for Task-Dependent Fine-Tuning of Pre-Trained Models: An Evolutionary Approach},
    year = {2024},
    publisher = {Association for Computing Machinery},
    volume = {4},
    number = {4},
    doi = {10.1145/3689827},
    journal = {ACM Trans. Evol. Learn. Optim.},
    articleno = {22},
    numpages = {23}
}

@article{jiang2021,
    author = {Jiang, Peng-Tao and Zhang, Chang-Bin and Hou, Qibin and Wei, Yunchao},
    year = {2021},
    month = {06},
    pages = {5875-5888},
    title = {LayerCAM: Exploring Hierarchical Class Activation Maps for Localization},
    volume = {30},
    journal = {IEEE Transactions on Image Processing},
    doi = {10.1109/TIP.2021.3089943}
}

@inproceedings{liu2022,
    author={Liu, Zhuang and Mao, Hanzi and Wu, Chao-Yuan and Feichtenhofer, Christoph and Darrell, Trevor and Xie, Saining},
    booktitle={2022 IEEE/CVF Conference on Computer Vision and Pattern Recognition (CVPR)}, 
    title={A ConvNet for the 2020s}, 
    year={2022},
    volume={},
    number={},
    pages={11966-11976},
    doi={10.1109/CVPR52688.2022.01167}
}

@article{radenovic2019,
    author={Radenović, Filip and Tolias, Giorgos and Chum, Ondřej},
    journal={IEEE Transactions on Pattern Analysis and Machine Intelligence}, 
    title={Fine-Tuning CNN Image Retrieval with No Human Annotation}, 
    year={2019},
    volume={41},
    number={7},
    pages={1655-1668},
    doi={10.1109/TPAMI.2018.2846566}
}

@inproceedings{pinaya2022,
    title={Brain imaging generation with latent diffusion models},
    author={Pinaya, Walter HL and Tudosiu, Petru-Daniel and Dafflon, Jessica and Da Costa, Pedro F and Fernandez, Virginia and Nachev, Parashkev and others},
    booktitle={MICCAI Workshop on Deep Generative Models},
    pages={117--126},
    year={2022},
    organization={Springer}
}

@inproceedings{loshchilov2019,
    title={Decoupled Weight Decay Regularization},
    author={Ilya Loshchilov and Frank Hutter},
    booktitle={International Conference on Learning Representations},
    year={2019},
    url={https://openreview.net/forum?id=Bkg6RiCqY7},
}

@inproceedings{paszke2019,
    author = {Paszke, Adam and Gross, Sam and Massa, Francisco and Lerer, Adam and Bradbury, James and Chanan, Gregory and others},
    booktitle = {Advances in Neural Information Processing Systems},
    title = {PyTorch: An Imperative Style, High-Performance Deep Learning Library},
    url = {https://proceedings.neurips.cc/paper_files/paper/2019/file/bdbca288fee7f92f2bfa9f7012727740-Paper.pdf},
    volume = {32},
    year = {2019}
}

@misc{cardoso2022,
    title={MONAI: An open-source framework for deep learning in healthcare}, 
    author={M. Jorge Cardoso and Wenqi Li and Richard Brown and Nic Ma and Eric Kerfoot and Yiheng Wang and others},
    year={2022},
    eprint={2211.02701},
    archivePrefix={arXiv},
    url={https://arxiv.org/abs/2211.02701},
}

@article{Ants2021,
    author = {Tustison, Nicholas J. and Cook, Philip A. and Holbrook, Andrew J. and Johnson, Hans J. and Muschelli, John and Devenyi, Gabriel A. and others},
    title = {The {ANTsX} ecosystem for quantitative biological and medical imaging},
    journal = {Scientific Reports},
    volume = {11},
    number = {1},
    pages = {9068},
    year = {2021},
    doi = {10.1038/s41598-021-87564-6}
}

@article{HdBet2019,
    author = {Isensee, Fabian and Schell, Marianne and Pflueger, Irada and Brugnara, Gianluca and Bonekamp, David and Neuberger, Ulf and others},
    title = {Automated brain extraction of multisequence MRI using artificial neural networks},
    journal = {Human Brain Mapping},
    volume = {40},
    number = {17},
    pages = {4952-4964},
    doi = {https://doi.org/10.1002/hbm.24750},
    year = {2019}
}

@article{WhiteStrip2014,
    title = {Statistical normalization techniques for magnetic resonance imaging},
    author = {Russell T. Shinohara and Elizabeth M. Sweeney and Jeff Goldsmith and Navid Shiee and Farrah J. Mateen and Peter A. Calabresi and others},
    journal = {NeuroImage: Clinical},
    volume = {6},
    pages = {9-19},
    year = {2014},
    issn = {2213-1582},
    doi = {https://doi.org/10.1016/j.nicl.2014.08.008}
}

@inproceedings{huang2017,
    author = {Huang, Gao and Liu, Zhuang and van der Maaten, Laurens and Weinberger, Kilian Q.},
    title = {Densely Connected Convolutional Networks},
    booktitle = {Proceedings of the IEEE Conference on Computer Vision and Pattern Recognition (CVPR)},
    year={2017},
    volume={},
    number={},
    pages={2261-2269},
    doi={10.1109/CVPR.2017.243}
}

@inproceedings{he2016,
    author={He, Kaiming and Zhang, Xiangyu and Ren, Shaoqing and Sun, Jian},
    booktitle={2016 IEEE Conference on Computer Vision and Pattern Recognition (CVPR)}, 
    title={Deep Residual Learning for Image Recognition}, 
    year={2016},
    volume={},
    number={},
    pages={770-778},
    doi={10.1109/CVPR.2016.90}
}

@article{johnson2019,
    title={Billion-scale similarity search with {GPUs}},
    author={Johnson, Jeff and Douze, Matthijs and J{\'e}gou, Herv{\'e}},
    journal={IEEE Transactions on Big Data},
    volume={7},
    number={3},
    pages={535--547},
    year={2019},
    publisher={IEEE}
}

@inproceedings{zeiler2014,
    author="Zeiler, Matthew D. and Fergus, Rob",
    title="Visualizing and Understanding Convolutional Networks",
    booktitle="Computer Vision -- ECCV 2014",
    year="2014",
    publisher="Springer International Publishing",
    address="Cham",
    pages="818--833"
}

@article{wang2021,
    author={Wang, Jingdong and Sun, Ke and Cheng, Tianheng and Jiang, Borui and Deng, Chaorui and Zhao, Yang and others},
    journal={IEEE Transactions on Pattern Analysis and Machine Intelligence}, 
    title={Deep High-Resolution Representation Learning for Visual Recognition}, 
    year={2021},
    volume={43},
    number={10},
    pages={3349-3364},
    doi={10.1109/TPAMI.2020.2983686}
}

@misc{abbas2023,
    title={SemDeDup: Data-efficient learning at web-scale through semantic deduplication}, 
    author={Amro Abbas and Kushal Tirumala and Dániel Simig and Surya Ganguli and Ari S. Morcos},
    year={2023},
    eprint={2303.09540},
    archivePrefix={arXiv},
    url={https://arxiv.org/abs/2303.09540}
}

@article{zhang2024,
    title={A Multimodal Biomedical Foundation Model Trained from Fifteen Million Image–Text Pairs},
    author={Sheng Zhang and Yanbo Xu and Naoto Usuyama and Hanwen Xu and Jaspreet Bagga and Robert Tinn and others},
    journal={NEJM AI},
    year={2024},
    volume={2},
    number={1},
    doi={10.1056/AIoa2400640}
}

@inproceedings{zhang2018,
    author = { Zhang, Richard and Isola, Phillip and Efros, Alexei A. and Shechtman, Eli and Wang, Oliver },
    booktitle = { 2018 IEEE/CVF Conference on Computer Vision and Pattern Recognition (CVPR) },
    title = {{ The Unreasonable Effectiveness of Deep Features as a Perceptual Metric }},
    year = {2018},
    volume = {},
    pages = {586-595},
    doi = {10.1109/CVPR.2018.00068},
    publisher = {IEEE Computer Society}
}

@inproceedings{radford2021,
  title={Learning transferable visual models from natural language supervision},
  author={Radford, Alec and Kim, Jong Wook and Hallacy, Chris and Ramesh, Aditya and Goh, Gabriel and Agarwal, Sandhini and others},
  booktitle={International conference on machine learning},
  pages={8748--8763},
  year={2021},
  organization={PmLR}
}

\end{document}